\documentclass[journal]{IEEEtran}
\usepackage{graphics} % for pdf, bitmapped graphics files
\usepackage{graphicx} 
\usepackage{epsfig} % for postscript graphics files
\usepackage{mathptmx} % assumes new font selection scheme installed
\usepackage{times} % assumes new font selection scheme installed
\usepackage{amsmath} % assumes amsmath package installed
\usepackage{amssymb}  % assumes amsmath package installed
\usepackage{cite}
\usepackage{algorithm}
\usepackage[noend]{algpseudocode}
\usepackage{verbatim}
\usepackage{commath}
\usepackage{bm}
\usepackage{booktabs}
\usepackage{multirow}
\usepackage{caption}
\usepackage{subcaption}
\usepackage{svg}

\usepackage{xcolor}

\DeclareMathOperator*{\argmin}{arg\,min}
\newcommand{\squishlist}{
\begin{list}{${\bullet}$}
	{ \setlength{\itemsep}{0pt}
		\setlength{\parsep}{3pt}
		\setlength{\topsep}{3pt}
		\setlength{\partopsep}{0pt}
		\setlength{\leftmargin}{1.5em}
		\setlength{\labelwidth}{1em}
		\setlength{\labelsep}{0.5em} } }
\newcommand{\squishend}{
\end{list}  
}

\usepackage[font={small,bf}]{caption}
\newcommand{\subparagraph}{}
\usepackage{titlesec}

\usepackage{etoolbox}
\usepackage{algorithm}
\usepackage[noend]{algpseudocode}

\makeatletter
\patchcmd{\ttlh@hang}{\parindent\z@}{\parindent\z@\leavevmode}{}{}
\patchcmd{\ttlh@hang}{\noindent}{}{}{}
\makeatother

\titlespacing{\section}{0pt}{*0}{*0}
\titlespacing{\subsection}{0pt}{*0}{*0}
\titlespacing{\subsubsection}{0pt}{*0}{*0}

\begin{document}
\title{A Method for Constraint Inference Using Pose and Wrench Measurements}
% Constraint inference using pose and wrench measurements
% pose or kinematics? 
\author{Guru ~Subramani,
        Michael ~Hagenow,
        Michael ~Gleicher,~\IEEEmembership{Member,~IEEE}
        and~Michael~Zinn,~\IEEEmembership{Member,~IEEE}% <-this % stops a space
\thanks{}% <-this % stops a space
\thanks{}% <-this % stops a space
\thanks{}}

%\markboth{IEEE Transactions on Robotics,~Vol.~x, No.~y, Month~yyyy}%
%{Shell \MakeLowercase{\textit{et al.}}: Bare Demo of IEEEtran.cls for IEEE Journals}
\maketitle

\begin{abstract}
Many physical tasks such as pulling out a drawer or wiping a table can be modeled with geometric constraints.
These geometric constraints are characterized by restrictions on kinematic trajectories and reaction wrenches (forces and moments) of objects under the influence of the constraint. 
This paper presents a method to infer geometric constraints involving unmodeled objects in human demonstrations using both kinematic and wrench measurements. 
Our approach takes a recording of a human demonstration and determines what constraints are present, when they occur, and their parameters (e.g. positions). By using both kinematic and wrench information, our methods are able to reliably identify a variety of constraint types, even if the constraints only exist for short durations within the demonstration.
We present a systematic approach to fitting arbitrary scleronomic constraint models to kinematic and wrench measurements.
Reaction forces are estimated from measurements by removing friction.    
Position, orientation, force, and moment error metrics are developed to provide systematic comparison between constraint models. 
By conducting a user study, we show that our methods can reliably identify constraints in realistic situations and confirm the value of including forces and moments in the model regression and selection process.

\end{abstract}


\IEEEpeerreviewmaketitle

\section{Introduction}
\IEEEPARstart{T}{asks} performed by robots frequently include constraints. For example, a manipulator is restricted to linear motion when pulling out a drawer. 
Representing physical interactions as constraints provides information about the permissible motion of objects in the environment and the permissible directions that forces and moments may be applied to them. 
This knowledge provides a way to inform the robot to restrict its movements to allowable configurations so it does not violate this closed kinematic chain (e.g., to plan motion consistent with the constraint \cite{Berenson2009,Stilman2007,Huh2018}) and to apply safe levels of force against the constraint to achieve the appropriate task-specific interaction (e.g., through hybrid control \cite{Raibert1981}).

Information about the constraints in the task is valuable to program a robot, but current approaches to determine constraint parameters are not straightforward. 
% If there were an easy source of constraint information, it would be useful for many robotics applications. 
For example, commercial robotic systems provide hybrid control abilities, however the constraints and their geometry must be determined manually. This makes such control difficult to use, limiting the ability of such systems to perform tasks involving constraints.
One solution is to infer constraints from a human demonstration, which eliminates the need for explicit programming \cite{Argall2009,PSMG19}. 
This paper presents an approach to infer constraints in human demonstrations of recorded force, moment, position, and orientation measurements.

\begin{figure}[t]
\centering
\includegraphics[width=3.2in]{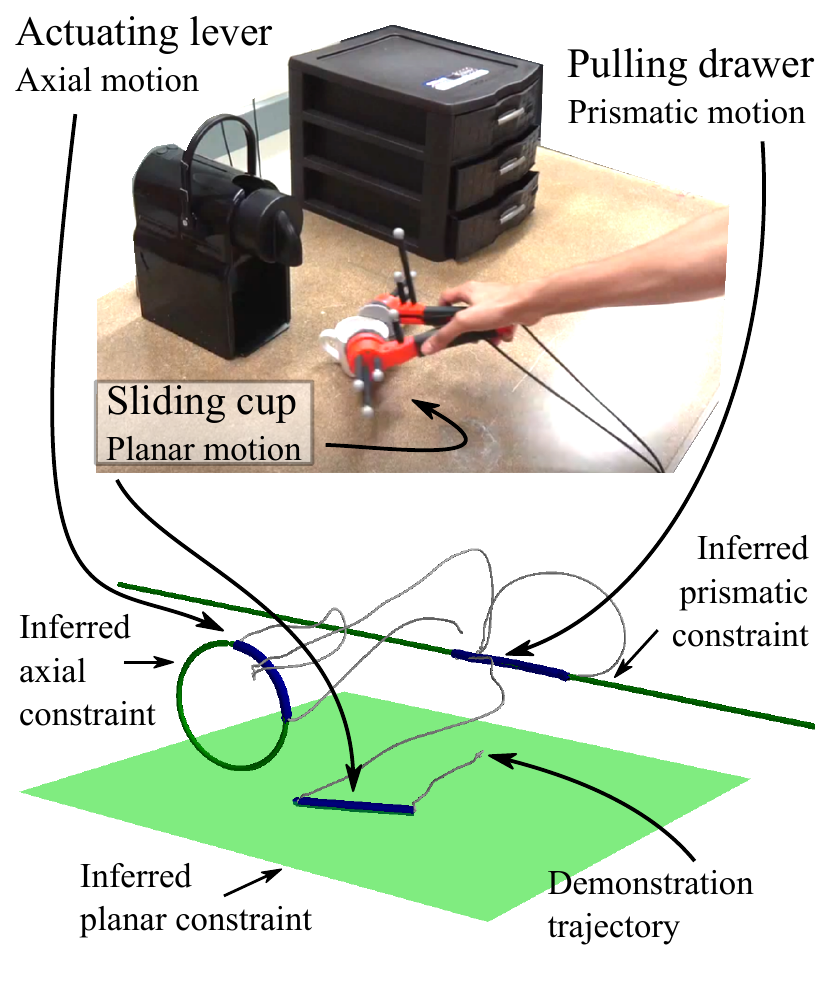}
\caption{Top: Demonstration of an espresso making task using instrumented tongs consisting of sliding an espresso cup, pulling out a drawer, and actuating an espresso lever.  Center: Inferred constraints in task. Our approach infers the samples associated with each constraint (segmentation), the type of constraint (model selection) and model parameters of each constraint (model regression) in the task.}
\label{fig:teaserEspressoTask}
\end{figure}

The requirements of a constraint inference system may vary dramatically depending on application. In manufacturing environments, the tool and environment geometry may be known beforehand, but other applications may require interacting with unknown objects. If a system is to perform inference using a natural demonstration (e.g the espresso making task in Fig. \ref{fig:teaserEspressoTask}), it may also need to handle demonstrations involving multiple constraint interactions, short constraint interactions, and other physical interactions such as picking up and placing objects.

Past work, summarized below, has shown that geometric constraint models can be fit from a demonstration containing sufficient data, however limited attention has been given to the ability to recognize constraints in a short demonstration. For example, the method described in \cite{Leopoldo2018} effectively learns a constraint model and develops a policy to act in the null-space of a constraint, which allows mapping of motion across constraints. To accomplish this, the constraint model is trained using 20 seconds of recorded demonstration data. Particularly when a task is performed in a natural way, there is no guarantee a model can be fit with the available duration of demonstration data. Certain applications might require a constraint inference method that minimizes the data required for fitting.

Representations of constraints can be general (i.e., a broad model that fits many types of constrained motion) \cite{Berenson2009,Li2017} or can be derived from geometric primitives \cite{arodriguez2008}. A general model offers flexibility to learn a wider range of constraints without specification of additional models, but is difficult to interpret. On the other hand, primitive models are more easily understood by human compatriots. For example, a rotational motion can be interpreted by an operator, whereas a general model or combination of low-level restrictions on rotational or translational motion may have little semantic meaning.
Great effort \cite{Dutre1996ContactEnergy,Joris1999,Lefebvre2003PolyhedralMotion,Meeussen2008} has been devoted to constraint fitting with known object geometries (e.g., polyhedral, cylindrical, and spherical tools) by combining kinematic and wrench relationships to fit the constraints of the interaction, including simultaneous active constraint interactions. Similarly, our method applies wrench-based fitting to our previous methods \cite{Subramani2018},\cite{SubramaniRAL2018} using nonlinear regression of constraint parameters, which allow for parameterization even when the interacting objects are not known.

The contribution of this paper is in presenting and assessing a constraint inference approach that uses both kinematic and wrench information to infer geometric constraints in a human demonstration, even if the objects are unknown.
Our approach can take a recording of a demonstration and determine what constraints are present, when they occur, and their parameters. It can reliably identify and localize a variety of constraint types, even when the constrained interactions are short. We build upon earlier work \cite{Subramani2018},\cite{SubramaniRAL2018} that uses only kinematics to fit constraint models involving unknown objects and uses forces and moment information for validation. More specifically, the new method extends our previous work by providing a systematic review of the value of wrench information throughout the constraint inference process, extending to new constraint models, and providing a more thorough and formalized explanation of the constraint inference approach. We provide new technical method details which are validated through new, extensive experimental results.

In our new method, we incorporate reaction forces and moments in an optimization objective to fit constraint models. We describe an approach to resolve reaction wrenches from wrench measurements which also contain constraint specific friction wrenches. Our method requires minimal assumptions about the friction model and does not require any specific friction parameters. We define a general set of position, force, orientation, and moment error metrics that are used to compare fit models with measured data. 
The product of these contributions is a constraint inference approach that explicitly recognizes the physical interaction that occurs when interacting with a constraint, where motions and wrenches are related in a manner dictated by the physics of the constrained interaction.  Our approach takes advantage of both wrench and kinematic measurements which enables improved performance over our previous methods that use only kinematics to fit constrained interactions with unknown objects. 
We assess our approach with a human subjects study with 9 participants that confirms both the approach's performance as well as the value of including wrench information directly into the nonlinear fitting of constraints, even if there are unknown objects.
\begin{figure}[t]
\centering
\includegraphics[width=3.2in]{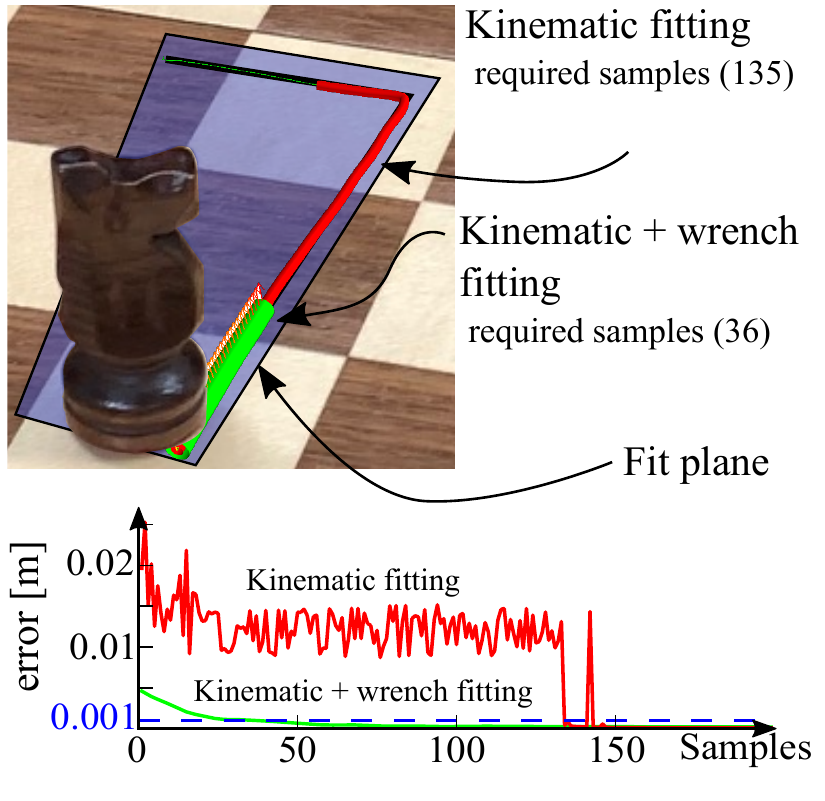}
\caption{A User demonstrates a chess move by sliding the knight against the chess board. Fitting the \emph{planar-relaxed constraint} using kinematic information [red] is compared against fitting with a combination of pose and wrench measurements (Kinematic-Wrench fitting) [green]. Kinematic-Wrench fitting is able to identify the plane using considerably fewer samples than kinematic fitting.}
\label{fig:teaser}
\end{figure}

\section{Constraint inference overview}
This paper provides a framework to infer geometric constraints in human demonstrations. 
Our approach is to first partition the human demonstration into constrained motion and unconstrained motion, fit all candidate constraint models to each constrained motion segment, and finally select the appropriate constraint models from the fit models. 
We organize the constraint inference as three sequential steps:

\begin{enumerate}
    \item \textbf{Demonstration segmentation:} Applications may require determining when constraints occur in a demonstration, especially when demonstrations represent a larger task.
    We discuss methods to segment the constrained motion from demonstrations in Section \ref{sec:segmentation} (after model regression and model selection to cover the more important contributions first). 
    Our approach takes advantage of the reaction forces and moments that exist during constraint interaction.
    \item \textbf{Model regression:} In Section \ref{sec:regression}, we show how nonlinear least squares regression can be used to fit geometric constraint models to kinematic and wrench measurements by leveraging the principle of virtual work.
    Our method derives reaction forces and moments from measured wrenches by considering contributing dynamics from friction. This is described in Section \ref{sec:frict}. We show that by simultaneously fitting with kinematic and wrench data, we are able to fit certain constraints using significantly less data compared to fitting with kinematic data (see Fig. \ref{fig:teaser}).
    \item \textbf{Model selection:} Once all models are fit to the constrained segments, each model fit must be evaluated and the appropriate model must be selected. 
    The regression objectives are inadequate to perform this comparison because they evaluate to different units and scales. 
    To address this problem, kinematic and wrench error metrics are derived for the general scleronomic constraint model to provide a consistent approach to evaluate models.  This is described in Section \ref{sec:modelselection}.
\end{enumerate}

\section{Mathematical Modeling of constraints} \label{sec:mathmodeling}
Our constraint inference approach requires defined mathematical representations of the constraints. 
We use models based on human interpretable geometric primitives to represent constrained motion such as the \emph{planar constraint} (e.g. a flat object wiping against a table), \emph{point-on-plane constraint} (e.g. the motion of a pen on paper), and \emph{axial rotation constraint} (e.g. the motion of a hinged door). 
These constraints are represented as mathematical equations that the tool or object must satisfy when under the influence of the constraint. The constraint models and relationship between the reaction wrenches (reaction forces and moments) imparted on constrained objects is described below. 

When a robot end effector or other object moves under the influence of a constraint, the degrees of freedom of its rigid body motion are restricted to a subset in SE(3). 
These restrictions may be represented as mathematical relationships between defined geometry on the body (e.g., a point on the body) and defined global geometry (e.g., a plane in the environment). 
For example, consider a pen moving against paper. The motion of the pen tip is constrained to the planar surface of the paper. In this example, the rigid body is the pen, the location of the pen tip is the defined geometry on the rigid body, and the planar surface represents the global geometry.  
Geometric constraint models consist of equations that define the possible configurations of the rigid body. 

Geometric constraints impose restrictions on the configurations on the rigid body. Applying forces and moments on a constrained rigid body creates reaction forces and moments between the rigid body and the constraint.  
The principle of virtual work expresses that reaction forces and moments from the constraint do not produce work. 
This property can be used to determine the permissible reaction forces and reaction moments of the constraint. 
It is also necessary to consider additional forces that may act on the rigid body such as friction, inertia, and gravity. These concepts are developed below.

\subsection{Geometric constraint models}
In our formulation, we consider scleronomic constraints, which are geometric constraint models that only impose restrictions on position and orientation (i.e. generalized coordinates of the rigid body) and do not vary with time\cite{goldstein2002classical}. In this paper we consider the eight constraint models shown in Fig. \ref{fig:constraints}. These constraints were chosen as examples of commonly occurring geometric primitives.  Our approach can be extended to other models satisfying the scleronomic constraint formulation. The constraint models considered here include:

\squishlist
\item \textbf{Point-on-line constraint:} The \emph{point-on-line constraint} is equivalent to drawing a line against a straight edge. A point on the rigid body is constrained to a line in space. 

\item \textbf{Planar constraint:} The \emph{planar constraint} is exemplified by an eraser moving against a whiteboard. The rigid body can only rotate about a vector perpendicular to the plane, and all points within the rigid body translate parallel to this plane. 

\item \textbf{Prismatic constraint:}
The \emph{prismatic constraint} represents translational motion in one direction. It is similar to pulling out a drawer. All points on the rigid body translate identically. 
%We assume the origin of the local coordinate frame of the body is contained within this line/axis.
\item \textbf{Axial rotation constraint} The \emph{axial rotation constraint} is similar to a door knob or a hinged door. All points on the rigid body rotate about an axis and translations are not permitted. This constraint is also referred to as a revolute joint.

\item \textbf{Relaxed constraints:} The \emph{planar}, \emph{axial}, and \emph{prismatic} constraints impose restrictions on the orientation of the rigid body. In certain cases, the demonstration tool may not rigidly fix the orientation of the manipulated rigid body. For example, when grabbing an object with a parallel gripper, there is a slip direction where the rigid body may rotate. The \emph{relaxed} versions of these constraints remove restrictions on the orientation of the constrained object in order to model this type of behavior.
\squishend

The detailed constraint mathematical models are developed in Appendix \ref{app:models}.  However, to provide context we present an extended description of the \emph{point-on-plane constraint} development in the following section.

\subsubsection{Extended Description for Point-on-plane Constraint} \label{sec:point_on_plane_model}

Consider a 6-degree of freedom rigid body of negligible inertial properties located in space through translational coordinates \textbf{\emph{r}} $\in \mathbb{R}^3$ and orientation coordinates represented by a unit quaternion \textbf{\emph{q}} $\in$ Spin(3). The measurement reference frame is the fixed frame on this rigid body.
The rigid body is constrained by $k$ constraint equations $\Phi$ : SE(3) $\Rightarrow$ $\mathbb{R}^k$:
\begin{equation}\label{eq:phi}
\Phi(\textbf{r},\textbf{q}) = 0
\end{equation}
The scleronomic constraint model represented by equation (\ref{eq:phi}) may be instantiated with a specific constraint model (e.g. with a \emph{point-on-plane constraint}) with associated model parameters $\boldsymbol{\alpha}$  which configure the constraint (e.g. the position and orientation of the plane). 
Thus, $\Phi$ becomes $\Phi(\textbf{r},\textbf{q},\boldsymbol{\alpha})$.

\begin{figure}[t]
\includegraphics[width=3.2in]{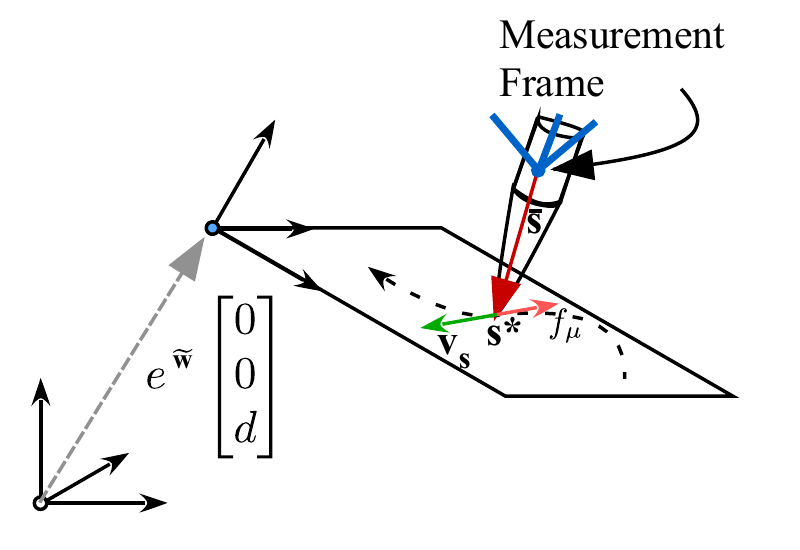}
\caption{\emph{Point-on-plane constraint} model. The point s* is constrained to the planar surface. $\bar{\mathbf{s}}$ is a fixed vector from the origin of the measurement frame to point s* (expressed in the measurement frame). The global x-y plane is transformed using the exponential map, ${e}^{\tilde{\mathbf{w}}}$, to create the planar surface.}
% \vspace*{-0.2in}
\label{fig:point_on_plane}
\end{figure}

The \emph{point-on-plane constraint} describes a point on the rigid body constrained to a plane in the environment such as a pencil tip (point) moving across paper (plane). An illustrated representation of this constraint is shown in Fig. \ref{fig:point_on_plane}.

\begin{figure*}[t]
\includegraphics[width=7.0in]{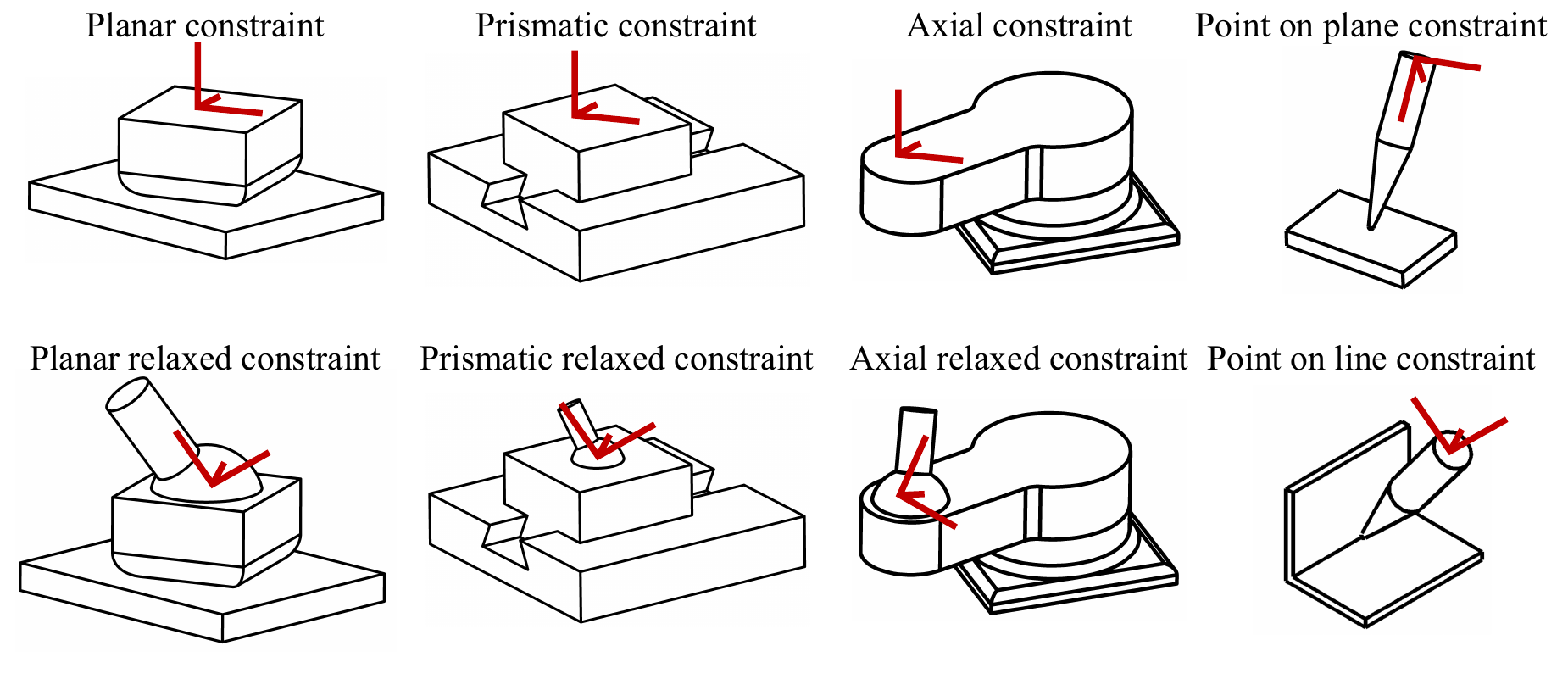}
\caption{Geometric constraints considered in this paper. The three constraint models \emph{planar constraint}, \emph{prismatic constraint} and \emph{axial constraint} enforce constraints on the orientation of the constrained object. The \emph{relaxed} counterparts are similar constraint models except with removed orientation constraints. The \emph{point-on-plane} and \emph{point-on-line} constraint models contain representations of the location of contact with respect to the measured position on the object. The measurement frame is shown in red. }
% \vspace*{-0.2in}
\label{fig:constraints}
\end{figure*}

A plane may be specified by applying a general displacement (i.e. translation and rotation) transformation of the \emph{x-y} plane which involves: 
\begin{enumerate}
\item translating the \emph{x-y} coordinate plane along the \emph{z}-axis
\item rotating the translated plane about the origin. 
\end{enumerate}
We represent the rotation transformation using two exponential coordinates $\textbf{w}=[{{w}_{x}},{{w}_{y}},0]\in \mathbb{R}^2$ corresponding to ${{e}^{{\tilde{\textbf{w}}}}}\in SO(3)$, the exponential map, which is equivalent to a rotation matrix with an axis of rotation in the \emph{x-y} plane. Rodrigues' rotation formula \cite{murray1994mathematical} is used to compute ${e}^{\tilde{\mathbf{w}}}$. $\tilde{\mathbf{w}}$ is the skew symmetric matrix similar to equation (\ref{eq:skewsym}) in Appendix \ref{app:ki}. The third term of \textbf{w} is zero because rotations about the \emph{z} - axis (perpendicular to the plane) do not alter the plane's geometry. The translation is represented by $d \in \mathbb{R}$. The normal vector on this plane is represented by ${e}^{\tilde{\mathbf{w}}}{\left[ 0\ \ 0\ \ 1 \right]}^{T}$ and the shifted origin of the \emph{x-y} plane is represented by ${{e}^{{\tilde{\mathbf{w}}}}}{{\left[ 0\ \ 0\ \ d \right]}^{T}}$.

A point $\mathbf{P}$ on the plane satisfies the following equation:
\begin{equation}
{{\left( {{e}^{{\tilde{\mathbf{w}}}}}{{\left[ 0\ \ 0\ \ d \right]}^{T}}-\mathbf{P} \right)}^{T}}{{e}^{{\tilde{\mathbf{w}}}}}{{\left[ 0\ \ 0\ \ 1 \right]}^{T}}=0
\end{equation}
\noindent which specifies that the dot product between a vector within the plane and the plane normal is zero. 

Consider a point $\mathbf{s^*} \in \mathbb{R}^3$ (in the global reference frame) defined as a point fixed to the rigid body and constrained to the plane. \textbf{s} is a vector directed from the origin of the measurement reference frame to the point \textbf{s*} (vector defined in the global reference frame). 
The local reference frame counterpart of \textbf{s} is $\overset{-}{\mathbf{s}}$ (whose values do not change with the motion of the measurement frame):
\begin{equation}
\mathbf{s^*}\equiv \mathbf{r}+\mathbf{s}\equiv \mathbf{r}+\textbf{A}(\mathbf{q})\overset{-}{\mathbf{s}}
\end{equation}
\noindent where $\mathbf{A}(\mathbf{q})$ is the orthogonal rotation matrix equivalent to the rotation represented by \textbf{q}.

The constraint is satisfied when $\mathbf{s^*}$ is constrained to the plane, i.e. equation $\Phi:\text{SE}(3) \Rightarrow \mathbb{R}^1$ is:
\begin{equation}\label{eq:pplanephi}
{{\Phi }_{1}}\equiv {{\left( {{e}^{{\tilde{\mathbf{w}}}}}{{\left[ 0\ \ 0\ \ d \right]}^{T}}-\mathbf{r}-\textbf{A}(\mathbf{q})\bar{\mathbf{s}} \right)}^{T}}{{e}^{{\tilde{\mathbf{w}}}}}{{\left[ 0\ \ 0\ \ 1 \right]}^{T}}=0
\end{equation}
\noindent 
The parameters of this constraint are $\boldsymbol{\alpha}  =  (\bar{\mathbf{s}},d,{{w}_{x}},{{w}_{y}})$. The derivation of the mathematical models of the other constraints shown in Fig. \ref{fig:constraints} is given in Appendix \ref{app:models}.

\subsection{Reaction forces and moments} \label{sec:genmod}
In addition to the constraint models, our method requires that we develop a relationship between reaction wrenches and the constraint models so that all measured data can be used in the constraint fitting process. The formulation used is standard in multi-body dynamics literature \cite{haug1989computer} and is provided here as a review.

We determine permissible reaction forces and moments using virtual displacements and the principle of virtual work. The first partial derivatives of constraint equations (\ref{eq:phi}) are given by:
\begin{equation}\label{eq:phiexp}
\delta \Phi ={{\Phi }_{\mathbf{r}}}\delta \mathbf{r}+{{\Phi }_{\mathbf{q}}}\delta \mathbf{q} = 0
\end{equation}
where ${{\Phi }_{\mathbf{r}}}$ and ${{\Phi }_{\mathbf{q}}}$ are the partial derivatives of (\ref{eq:phi}) and $(\delta \mathbf{r},\delta \mathbf{q})$ are the virtual displacements.
Equation (\ref{eq:phiexp}) may also be written using virtual rotation variable $\delta \boldsymbol{\pi} $ \cite{haug1989computer}. The virtual rotation variable is related to the angular velocity of a rigid body. This relationship is similar to how the virtual displacement $\delta \mathbf{p}$ is related to linear velocity. 
\begin{equation}\label{eq:virt}
\delta \Phi ={{\Phi }_{\mathbf{r}}}\delta \mathbf{r}+{{\Phi }_{\boldsymbol{\pi} }}\delta \boldsymbol{\pi} = 0
\end{equation}
Equation (\ref{eq:virt}) is similar to the Pfaffian constraint form used in the robotics literature \cite{choset2005principles}.
The permissible reaction forces $\mathbf{F}_r\in {{\mathbb{R}}^{3}}$ and reaction moments $\mathbf{N}_r\in {{\mathbb{R}}^{3}}$ that the constraint applies to the constrained body must satisfy the virtual work equation \cite{lanczos2012variational} because constraint reaction forces and moments do not produce work. The principle of virtual work requires:
\begin{equation} \label{eq:vw}
\delta {{\mathbf{r}}^{T}}\mathbf{F}_r+\delta {{\boldsymbol{\pi} }^{T}}\mathbf{N}_r=0
\end{equation}
Combining equations (\ref{eq:virt}) and (\ref{eq:vw}) using the Lagrange multiplier theorem \cite{haug1989computer}:
\begin{gather}
{{\Phi }_{\mathbf{r}}}^{T}\boldsymbol{\lambda}+\mathbf{F}_r=0 \label{eq:vwf}\\
\Phi _{\boldsymbol{\pi} }^{T}\boldsymbol{\lambda}+\mathbf{N}_r=0 \label{eq:vwm}
\end{gather}
where $\mathbf{\boldsymbol{\lambda}} \in {{\mathbb{R}}^{k}}$ are the Lagrange multipliers for the $k$ constraint equations. Equations (\ref{eq:vwf}) and (\ref{eq:vwm}) provide a relationship between the constraint equations $\Phi$ and the permissible reaction forces and moments.
${{\Phi }_{\boldsymbol{\pi} }}\delta \boldsymbol{\pi} $ is computed from ${{\Phi }_{\mathbf{q}}}\delta \mathbf{q} $. This is provided in Appendix \ref{app:ki}. 

\subsection{Resolving reaction forces and moments from wrench measurements}\label{sec:frict}
Equations (\ref{eq:vwf}) and (\ref{eq:vwm}) relate the reaction forces and moments with the constraint equations described by equation (\ref{eq:phi}). One challenge is that reaction forces and moments are not measured directly. Interactions with the physical environment involve additional forces such as friction, inertia, gravity, forces from object compliance, and forces from task-specific impedances. To fit the constraint models, we need to estimate the reaction wrenches from measurement wrenches containing these additional effects.

To do so, we assume quasi-static motion and choose to model only friction in our analysis of constrained motion. 
In cases where demonstrations are highly dynamic, such as involving objects with large inertia and acceleration, this approach may not be directly applicable and such dynamic effects must be modeled and incorporated appropriately. 

Quasi-static motion against constraints generates frictional forces due to the sliding contact between surfaces. 
Forces attributed to friction oppose the motion of the constrained object.
For example, when writing on paper (\emph{point-on-plane constraint}), friction force opposes the motion of the pen tip. 
Friction forces may manifest themselves differently for each constraint and there is no general way to describe them.

However, friction forces and moments can be modeled for each constraint with certain assumptions. Our basic intuition is to remove friction in the direction of motion of objects in contact.
We derive a relationship between the reaction forces and moments with the measured forces and moments as functions of the parameters of the constraint:
\begin{gather}
    \mathbf{F}_{r} =  \mathbf{F_r}(\mathbf{r},\mathbf{q},\mathbf{v},\boldsymbol{\omega},\mathbf{F},\mathbf{N},\boldsymbol{\alpha}) \label{eq:frictionf}\\
    \mathbf{N}_{r} =  \mathbf{N_r}(\mathbf{r},\mathbf{q},\mathbf{v},\boldsymbol{\omega},\mathbf{F},\mathbf{N},\boldsymbol{\alpha}) \label{eq:frictionn}
\end{gather}

\noindent where the measured quantities are forces ($\mathbf{F}$), moments ($\mathbf{N}$), linear velocity ($\mathbf{v}$), and  angular velocity ($\boldsymbol{\omega}$) and the parameters of the constraint are $\boldsymbol{\alpha}$. 
The specific formulation of equations (\ref{eq:frictionf}) and (\ref{eq:frictionn}) for each constraint model is provided in Appendix \ref{app:models}. The resolving of reaction forces and moments for the \emph{point-on-plane} constraint is provided here as an example. 

\subsubsection{Resolving reaction forces and moments for the point-on-plane constraint}
The goal is to resolve the reaction wrenches from the measured wrenches (i.e. determine expressions for $\mathbf{F_r}(\mathbf{r},\mathbf{q},\mathbf{v},\boldsymbol{\omega},\mathbf{F},\mathbf{N},\boldsymbol{\alpha})$ and $ \mathbf{N_r}(\mathbf{r},\mathbf{q},\mathbf{v},\boldsymbol{\omega},\mathbf{F},\mathbf{N},\boldsymbol{\alpha})$).
We assume that inertia of the constrained object is negligible and the only wrenches acting on the object are friction and reaction wrenches.

The force and moment balance equation for the rigid body at the measurement frame are as follows: 
\begin{gather}
    \mathbf{F} = \mathbf{F}_{\mu} + \mathbf{F_r} \label{eq:forcebalance}\\
    \mathbf{N} = \mathbf{N}_{\mu} + \mathbf{N_r} \label{eq:momentbalance}
\end{gather}

\noindent where $\mathbf{F}$ is the measured force, $\mathbf{F}_{\mu}$ is the friction force, $\mathbf{F_r}$ is the reaction force, $\mathbf{N}$ is the measured moment, $\mathbf{N}_{\mu}$ is the friction moment, and $\mathbf{N_r}$ is the reaction moment acting at the measurement frame. The measured force and moments are expressed in the global frame (i.e., rotated from the original force torque sensor measurements).

For the \emph{point on plane constraint} seen in Fig. \ref{fig:point_on_plane}, point $\mathbf{s^*}$ on the rigid body is in contact with the plane. $\mathbf{s^*}$ moves with velocity $\mathbf{v}_{\mathbf{s}}$. 
%% MG - need The
The friction %% Friction  
force acts at point $\mathbf{s^*}$ on the rigid body and is directed along $\mathbf{v}_{\mathbf{s}}$ ($\widehat{\mathbf{v}_{\mathbf{s}}}$ is a unit vector in this direction). Any force in the direction of $\mathbf{v}_{\mathbf{s}}$ is assumed to be 
%% MG - not sure why manual link breaks in paragraph
friction. 
%\noindent 
$\mathbf{F}_{\mu}$ is the friction force located at point $\mathbf{s^*}$ directed along $\mathbf{v}_{\mathbf{s}}$: 
\begin{equation}\label{eq:frictionforce}
    \mathbf{F}_{\mu} = (\mathbf{F} \cdot\widehat{\mathbf{v}_{\mathbf{s}}})\widehat{\mathbf{v}_{\mathbf{s}}}
\end{equation}
\noindent where $\mathbf{v}_\mathbf{s}$ is given by: 
\begin{equation}\label{eq:velocity direction}
    \mathbf{v}_\mathbf{s} = \mathbf{v} + \boldsymbol{\omega} \times \mathbf{s}
\end{equation}
\noindent and $\mathbf{s}$ is the vector from the measurement frame to point $\mathbf{s^*}$ in the global frame. 

We assume that at $\mathbf{s^*}$, no moment is transferred (including friction moment) between the rigid body and the plane. The moment induced by the friction forces at $\mathbf{s^*}$ is given by: 
\begin{equation} \label{eq:frictionmoment}
    \mathbf{N}_{\mu} = \mathbf{s} \times \mathbf{F}_{\mu}
\end{equation}

Combining equations (\ref{eq:forcebalance}) through (\ref{eq:frictionmoment}) results in expressions for the reaction forces and moments as a function of measured values: 
\begin{gather}
    \mathbf{F_r} = \mathbf{F} - (\mathbf{F} \cdot\widehat{\mathbf{v}_{\mathbf{s}}})\widehat{\mathbf{v}_{\mathbf{s}}}\\
    \mathbf{N_r} = \mathbf{N} - \mathbf{s} \times (\mathbf{F} \cdot\widehat{\mathbf{v}_{\mathbf{s}}})\widehat{\mathbf{v}_{\mathbf{s}}}
\end{gather}

With equations to resolve reaction forces and moments, Equations \ref{eq:vwf} and \ref{eq:vwm} can be evaluated using measured data.  Similar mathematical relationships between measured forces and moments and reaction forces and moments for the other constraint models are given in Appendix \ref{app:models}.

%
% For example, friction forces are applied at the pen tip offset from the measurement frame. This causes a friction moment at the measurement frame. However, in the case of a whiteboad eraser moving against a whiteboard friction forces do not influence the .
% While it may appear that friction wrenches may only occur in the direction of motion, this is not true. This is because the virtual work equation (\ref{eq:vw}) which relates reaction forces and moments with linear and angular velocity are coupled. It is possible for reaction forces and moments to have components in the direction of motion. For example, it is possible to apply a moment on a door and rotate it while experiencing reaction forces along the direction of motion. (See figure \ref{fig:rotreact}.) 

% \begin{figure}[h]
% \includegraphics[width=3.2in]{./figures/door_rotate.png}
% \caption{Reaction forces and moments may be directed along the direction of motion. A hinge is depicted with applied forces and moments (green) causing a reaction force in the direction of motion.}
% \label{fig:rotreact}
% \end{figure}

\section{Least squares regression of constraint models} \label{sec:regression}
Using the constraint and reaction wrench equations developed in the previous sections, the constraint model parameters can be fit to the measured data. The generalized constraint equations given in (\ref{eq:phi}) are used to formulate a kinematic error objective  
%The kinematic error objective uses the recorded positions and orientations of the rigid body to fit the model. 
while equations (\ref{eq:vwf}) and (\ref{eq:vwm}) are used to formulate a wrench error objective. 
The two error objectives are cooperative and are included in a summed objective to fit recorded demonstration data to a constraint model. Let the parameters of the constraint be $\boldsymbol{\alpha}$. 
 and let the $i^{th}$ sample of data be represented with a tuple $\mathbf{p}_i = (\mathbf{r}_i, \mathbf{q}_i, \mathbf{F}_i, \mathbf{N}_i, \mathbf{v}_i, \boldsymbol{\omega_i})$ which consists of measured poses, wrenches, and velocities estimated from the kinematic measurements. The error objectives are defined across the N samples of data $\mathbf{p} := \{\mathbf{p}_1, ..., \mathbf{p}_N\}$. The kinematic and wrench error objectives are described below.

\subsection{Kinematic error objective}
The kinematic error objective is developed to incorporate kinematic measurements during fitting and is given by: 

\begin{equation}\label{eq:K_error}
K_{error}(\boldsymbol{\alpha},\mathbf{p},\mathbf{w}) = \sum\limits_{i = 1}^{N}{w_i\|\Phi(\mathbf{p}_i,\boldsymbol{\alpha})\|}
\end{equation}
\noindent where $\|\Box\|$ is the 
L(2) %% l2 
norm and $\mathbf{w} := \{w_1, ..., w_N\}$ are the weights on each sample. 
Sample weights allow increasing the influence of certain samples over others. Some samples included in the regression may not be part of the constraint or may be caused by measurement errors and should have less influence when fitting. 
For the kinematic error objective, only $(\mathbf{r},\mathbf{q})$ is required from tuple $\mathbf{p}$.

\subsection{Wrench error objective}
\label{sec:wrenchobjective}
A wrench error objective is developed to incorporate wrench measurements during fitting. 
Equations (\ref{eq:vwf}) and (\ref{eq:vwm}) provide a relationship between the constraint equations and the applied forces and moments in a demonstration.
To apply equations (\ref{eq:vwf}) and (\ref{eq:vwm}) for regression, 
the Lagrange multipliers must be determined jointly with the parameters of the constraint. 
This may be formulated as a weighted least squares objective:  

\begin{equation}\label{eq:lag}
W_{error}(\boldsymbol{\alpha},\mathbf{p},\mathbf{w}) = \sum\limits_{i = 1}^{N}w_i \min_{\lambda} 
      \left( \|\Phi _{\mathbf{r}}^{T}\boldsymbol{\lambda}+\mathbf{F}_{\mathbf{r}i}\|
 +\|\Phi _{\boldsymbol{\pi} }^{T}\boldsymbol{\lambda}+\mathbf{N}_{\mathbf{r}i}\| \right)
\end{equation}

Equations (\ref{eq:vwf}) and (\ref{eq:vwm}) are linear in $\boldsymbol{\lambda}$ and may be solved using linear least squares regression. The least squares error of $\mathbf{A}\boldsymbol{\lambda} = \mathbf{b}$ is denoted as $LSQ(\mathbf{A},\mathbf{b})$ where $\mathbf{A}$ and $\mathbf{b}$ are given by:

\begin{equation}
\mathbf{A} = 
\begin{bmatrix}
\Phi _{\mathbf{r}}^{T}\\
\Phi _{\boldsymbol{\pi}}^{T}
\end{bmatrix}
\end{equation}
and 
\begin{equation}
\mathbf{b} = 
\begin{bmatrix}\label{eq:blinls}
-\mathbf{F}_\mathbf{r}\\
-\mathbf{N}_\mathbf{r}
\end{bmatrix}
\end{equation}
where the reaction forces and moments, $\mathbf{F_r}$ and $\mathbf{N_r}$, are evaluated from measured forces and moments using equations (\ref{eq:frictionf}) and (\ref{eq:frictionn}).

Equation (\ref{eq:lag}) may be reformulated as: 

\begin{equation}\label{eq:W_error}
W_{error}(\boldsymbol{\alpha},\mathbf{p},\mathbf{w}) = \sum\limits_{i = 1}^{N} w_i LSQ\left(
\begin{bmatrix}
\Phi _{\mathbf{r}}^{T}(\mathbf{p}_i,\boldsymbol{\alpha})\\
\Phi _{\boldsymbol{\pi}}^{T}(\mathbf{p}_i, \boldsymbol{\alpha})
\end{bmatrix},\begin{bmatrix}
-\mathbf{F}_{\mathbf{r}i}\\
-\mathbf{N}_{\mathbf{r}i}
\end{bmatrix}\right) 
\end{equation}

The reaction forces and reaction moments are jointly estimated by including their expressions in the optimization objective.

\subsection{Maximum likelihood estimate}
Using the kinematic and wrench error objective functions we can determine the most likely model parameters $\boldsymbol{\alpha}$ given model $\Phi$ and demonstrator data $\mathbf{p}$.
The maximum likelihood estimate of model parameters $\boldsymbol{\alpha}$ based on observations $\mathbf{p}$ is $\hat{\boldsymbol{\alpha}}_{MLE}$ and can be estimated as:
\begin{equation}\label{eq:overall}
    \hat{\boldsymbol{\alpha}}_{MLE} = \argmin_{\boldsymbol{\alpha}} \left( W_{error}(\boldsymbol{\alpha},\mathbf{p},\mathbf{w}) + K_{error}(\boldsymbol{\alpha},\mathbf{p},\mathbf{w})\right)
\end{equation}
Since the two objectives $W_{error}$ and $K_{error}$ are physically related through virtual work, they behave cooperatively. Experimental evaluation indicates that weighting the two objectives has little effect on the overall fit. Equation (\ref{eq:overall}) can be used to estimate the constraint parameter values.

\subsection{Robust regression}\label{sec:robustregression}
Segmentation error and non-gaussian noise from sensor measurements may cause the nonlinear least squares optimization in equation (\ref{eq:overall}) to fit poorly. All of the samples may not be part of the constraint. To address these issues, we use iterative reweighted least squares (IRLS) \cite{burrus2012iterative} to estimate the sample weights associated with the constraint jointly with the model parameters which helps to reject outliers. 

The approach initializes equal weight values to each sample. It performs an optimization using a standard non-linear optimization algorithm (in this work we use BFGS \cite{fletcher2013practical}) using these weights. It calculates the errors associated with each sample and down weights samples that have large error. 
A re-weighting step is used to normalize the weights using a regularization function to prevent aggressive down-weighting of samples so that all samples continue to have an influence on the objective.
Using these new weighted values, the algorithm performs another iteration of optimization. 

The process of finding the most likely value of $\boldsymbol{\alpha}$ is given in Algorithm \ref{alg:robust}. An error convergence exit criteria is not necessarily applicable since all models are fit to each constraint segment. In this work, we choose a basic exit criteria of maximum number of iterations N. A more advanced criteria could be considered in future implementations.

\begin{algorithm}
\caption{Iterative reweighted least squares regression}\label{alg:robust}
\begin{algorithmic}[1]
\Procedure{RobustRegression}{\textbf{X}, \textbf{w}, N}
\State initialize(\textbf{w}) \Comment{set all weights to 1}
\For{j = 1 ... N} \Comment number of iterations
\State \textbf{w}$\gets$ \textbf{w} / sum (\textbf{w}) \Comment{Normalization}
\State $\boldsymbol{\alpha} \gets$ FitFunction(\textbf{X},\textbf{w}) \Comment{BFGS minimization}
\State \textbf{E}$\gets$ ErrorFunction(\textbf{X},$\boldsymbol{\alpha}$,\textbf{w}) \Comment{vector of fit errors}
\State \textbf{w} $\gets$ $1/(1 + \text{E}_i^{0.1})$ \Comment{Regularization}
\EndFor
\State \textbf{return} $\boldsymbol{\alpha}$\Comment{}
\EndProcedure
\end{algorithmic}
\end{algorithm}

\begin{figure}[h]\
\centering
\includegraphics[width=3.2in]{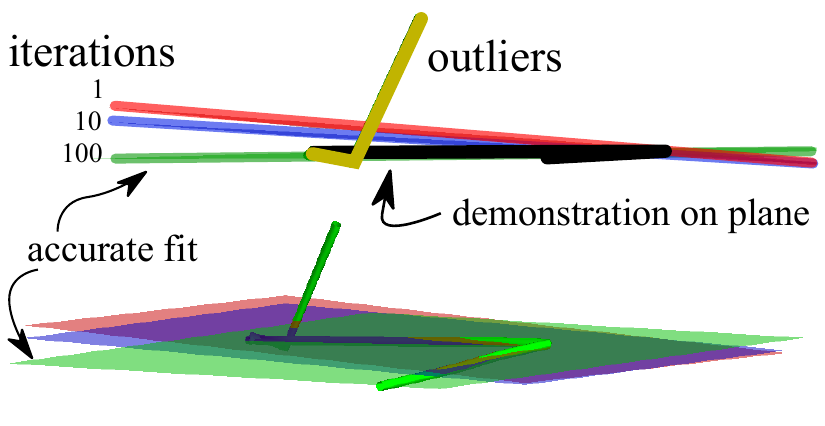}
\caption{Bottom: the perspective view of the demonstration in Fig. \ref{fig:teaser} of sliding the chess piece across the board (\emph{planar constraint}) is shown. The demonstration is intentionally segmented incorrectly [yellow outliers]. Iterative reweighted least squares is used to robustly fit the plane in the demonstration. 1 iteration [red] is equivalent to not using a robust fitting algorithm. This leads to a plane that has incorrect orientation. 100 iterations of IRLS [green] accurately fits the model to the constraint. }
\label{fig:irls}
\end{figure}

To illustrate the benefit of the iterative reweighted least squares we consider the example of moving a chess piece across a chessboard (shown in Fig. \ref{fig:teaser}). In this case, we intentionally segment the demonstration incorrectly so a portion of the demonstration is in free space. When fitting the model without iterative reweighted least squares (IRLS) regression, the fit plane is offset due to outliers caused by incorrect segmentation. After 100 iterations of IRLS the fit converges to the true constraint model (see Fig. \ref{fig:irls}). With IRLS, we can estimate the constraint parameters from imperfect measurement data.  

\section{Geometric constraint model selection} \label{sec:modelselection}
Once the parameters have been fit for a set of candidate models, it is necessary to identify the model most consistent with the measured data. The difficulty of the model selection problem is a function of the number of candidate models and the uniqueness of these models.

To simplify the model comparison process, we create a single selection metric that can be used across models. In our formulation, the optimization fit error is not appropriate as the models consist of regressor functions that are not guaranteed to have consistent units. To allow for comparison, kinematic and wrench error metrics are defined in terms of measured quantities (m, rad, N, Nm). These metrics are compared to predefined thresholds and are combined into a single error metric by counting measurements that exceed these thresholds. We refer to this process as error tallying.

\subsection{Error metrics}
Position and force information allow for the construction of separate kinematic and wrench error metrics.
An error metric is defined as a measure of distance between a measurement $(e.g. \mathbf{r},\mathbf{q},\mathbf{F},\mathbf{N})$ and the constraint model.
\subsubsection{Kinematic error metrics}
Given that a constraint can enforce both a position and an orientation, there is a need for two kinematic metrics. A position distance error metric, $D_{\mathbf{r}}(\Phi(\mathbf{r}^*,\mathbf{q}^*),\mathbf{r},\mathbf{q})$, and an orientation distance error metric, $D_{\mathbf{q}}(\Phi(\mathbf{r}^*,\mathbf{q}^*),\mathbf{r},\mathbf{q})$, are defined for a pose measurement $(\mathbf{r},\mathbf{q}) \in SE(3)$ where $(\mathbf{r}^*,\mathbf{q}^*)$ is coincident with the constraint (satisfies $\Phi$). Evaluation of $(\mathbf{r}^*,\mathbf{q}^*)$ consists of finding the point on the constraint that is closest in overall distance to the measured point. Determining these values is posed as a least squares regression where $(\mathbf{r}^*,\mathbf{q}^*)$ are solved simultaneously:

\begin{equation}
\hat{\mathbf{r}}^*,\hat{\mathbf{q}}^* = \argmin_{\mathbf{r}^*,\mathbf{q}^*}{}\left(
\begin{aligned}
       &||\mathbf{r}^* - \mathbf{r}|| + \angle\left(conj(\mathbf{q})\star \mathbf{q}^*\right)\\
 +\quad& W\Phi(\mathbf{r}^*,\mathbf{q}^*)^T \Phi(\mathbf{r}^*,\mathbf{q}^*) \\
 +\quad& W||\mathbf{q}^{*T} \mathbf{q}^* - 1||_2 
\end{aligned}
\right)
\end{equation}
\noindent where $W$ is a large weight (1000) on the constraint equations to ensure that $\mathbf{r}^*$ and $\mathbf{q}^*$ are fixed to the constraint. $\angle$ is the rotation angle of a quaternion, $\star$ represents quaternion multiplication, and $conj$ represents the conjugate of a quaternion. $\angle \left(conj(\mathbf{q})\star \mathbf{q}^*\right)$ computes the angle between q and q* (bounded to 180 degrees). While the position and orientation have different units and bounds, we did not add any scaling or normalization factors as for many of the constraints, the position and orientation optimize independent aspects of the constraint.

The kinematic error metrics, $D_{\mathbf{r}}$ and $D_{\mathbf{q}}$ can subsequently be computed as:
\begin{equation}
\label{eq:kerrormetricr}
D_{\mathbf{r}} =  ||\hat{\mathbf{r}}^\star - \mathbf{r}||
\end{equation}
\begin{equation}
\label{eq:kerrormetricq}
D_{\mathbf{q}} =  \angle\left(conj(\mathbf{q})\star \hat{\mathbf{q}}^\star\right)\space
\end{equation}
%where $\star$ represents quaternion multiplication.
\subsubsection{Wrench error metrics}
Measured forces and moments provide metrics that quantify inconsistency with the constraint model, $\Phi$. By computing the Lagrange multipliers individually for each measurement (see Section \ref{sec:wrenchobjective}), a force and moment error can be written as the residual of the virtual work equations previously defined in equations (\ref{eq:vwf}) and (\ref{eq:vwm}). The values are normalized with respect to the measured forces and moments to remove a dependence on magnitude, resulting in the wrench error metrics, $E_f$ and $E_n$, associated with the measured forces and measured moments respectively :

\begin{equation}
\label{eq:werrormetricr}
E_{\mathbf{F}} =   ||\Phi _{\mathbf{r}}^{T}\lambda+\mathbf{F}|| / ||\mathbf{F}|| 
\end{equation}
\begin{equation}
\label{eq:werrormetricq}
E_{\mathbf{N}} =   ||\Phi _{\boldsymbol{\pi} }^{T}\lambda+\mathbf{N}|| / ||\mathbf{N}||\space\\ 
\end{equation}

\subsubsection{Combined error metric}
For model selection, we desire a single metric capable of disambiguation between candidate models. A weighted sum of the individual error metrics would contain inherent bias dependent on the range of motion and level of applied forces and moments. To circumvent this, we introduce a combined error metric (CE) which is a function of the number of samples of the constrained motion that exceed predefined error thresholds given as $r_{SE}$, $q_{SE}$, $f_{SE}$, and $n_{SE}$. The combined metric considers all measured variables and is normalized such that models can be compared directly. The error thresholds, which have units corresponding to the units of the corresponding error metric, are determined by evaluating the error metrics using the numerical values of the measurement error, which is a function of the experimental setup.  The combined error metric (summed over the measured samples, $n$) is evaluated as:
\begin{equation}\label{eq:MS}
CE=\sum\limits^{N} \left(
\begin{aligned}
       &(D_r(n)>r_{SE}) \\
       &+ (D_q(n)>q_{SE}) \\
       &+ (E_f(n)>f_{SE})\\
       &+ (E_n(n)>n_{SE}) \\
\end{aligned}
\right)+1
\end{equation}

Section \ref{sec:modelselectionexp} describes how the thresholds are set for each of the error metrics based on an estimate of the experimental measurement accuracy. Our combined error metric provides one method that allows combination of the different error metrics, but does require thresholds based on an estimate of system error. Future work could include a systematic comparison between this method and other methods for combining the various forms of information, including a median error-based metric or an energy-based formulation.

For each model in the set of models, i.e., $m\in{M}$, a likelihood metric, $\hat{L}_m$, is evaluated from the combined error metric:

\begin{equation}
    \hat{L}_m =  \frac{CE(m)^{-1}}{\sum\limits_{m}^{M}{CE(m)^{-1}}}
    \end{equation}

Note, the initial value of the combined error metric, $CE$, as defined in equation (\ref{eq:MS}) is set equal to one to avoid division by zero in the calculation of the likelihood metric.

\section{Segmentation}\label{sec:segmentation}
A demonstration may contain a single constraint or multiple constraints of various types.  Additionally, an individual constraint may be visited multiple times during a single demonstration.  Finally, free space motion may occur between constraint interactions or different constraint interactions may occur consecutively, with no intermediate free-space motion. 

In order to apply regression and model selection to each constraint interaction individually, the associated samples of the constraint must be determined from the demonstration. 
While user-specified segmentation is possible, it is desirable to make the process automatic to avoid the need for user intervention to define constraint boundaries for segmentation.  

Prior work has explored the segmentation problem.
One approach \cite{SubramaniRAL2018} is to fit each constraint type over a small sliding window. 
Similar constraint fits are then clustered over the entire demonstration.  While effective, this approach is computationally expensive and limited to simple constraint models. It also requires significant tuning of heuristics by the user to achieve reliable segmentation.  
In another method, \cite{Niekum2015}, change point detection is used to determine constraint locations and can detect transitions between constraints without free-space motion, but this method is limited to simple models such as lines and arcs.  

In our method, we use an extension to the basic idea of force thresholding. If a measured force exceeds a specified threshold (e.g. twice the measurement noise of the system), then a constraint may be present. 
The result is a Boolean list of contiguous samples indicating possible constrained motion segments and free space motion segments. One limitation of this simple segmentation method is that free space motion must occur between each constraint interaction. While our experimental demonstrations are designed to meet this assumption, other methods that identify transitions between contiguous constraints or that incorporate multiple constraints within a single model could be incorporated in the future.   

Since measured forces can occur due to other effects including gravitational force or forces induced from accelerating motion, force thresholding is susceptible to false positive segments. To compensate for this we extend force thresholding to include a constraint null hypothesis. 

\subsection{Constraint Null Hypothesis} \label{sec:nullhyp}
The goal of the constraint null hypothesis is to remove false positive constraint fits. The fitting and model selection procedures described in sections \ref{sec:regression} and \ref{sec:modelselection} determine the best model that matches the data for a particular segment. For each of these segments, we evaluate whether the constraint should be kept or discarded (i.e. is it a false positive). An error threshold for each error metric (position, orientation, force and moment) is applied to each sample in the segment and contributes to cumulative tallies $V_r$, $V_q$, $V_f$ and $V_n$. A tally is awarded when the error is more than the threshold.
The thresholds are set above the system's measurement noise.
If the percentage of cumulative tallies for any metric is greater than an acceptable maximum percentage (e.g. 75\% in our implementation), the segment is discarded. Additionally, any segment consisting of only a few samples will easily fit a given model, but does not lend to confidence that the data is actually constrained.
For example, if the segment is only two points, the \emph{prismatic relaxed} model will always fit with zero error.
To avoid this case, we discard segments of very short duration (in our implementation, less than or equal to five samples).

\begin{figure}[t]
\centering
\includegraphics[width=3.2in]{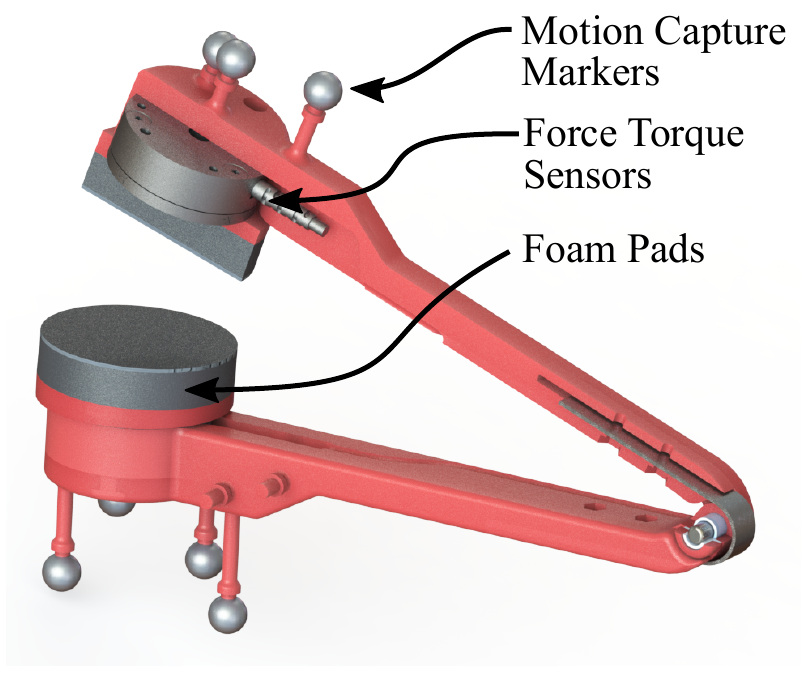}
\caption{Instrumented tongs - a handheld device used to record human demonstrations.}
\label{fig:tongs}
\end{figure}

\begin{figure*}[t]
\centering
\includegraphics[width=7in]{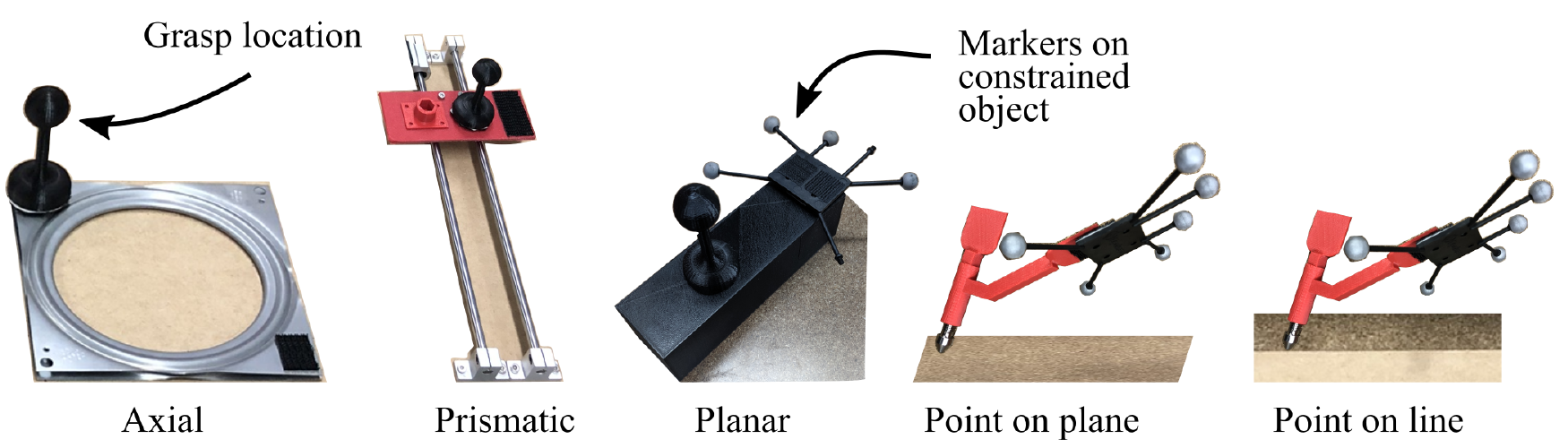}
\caption{Physical constraints used in experiment.}
\label{fig:testbed}
\end{figure*}

\section{Experimental evaluation and discussion}
\label{sec:expeval}
The constraint inference approach is evaluated in three experiments.  In the first two experiments, we desire to assess the situations in which wrench information improves constraint fitting of unknown objects. To do so, we compare a method using pose and wrench measurements to an approach that uses only pose measurements. These two experiments include (1) an evaluation of \emph{model fitting} (Section \ref{sec:fittingexp}) for demonstrations containing a single constraint and (2) an evaluation of \emph{model selection} (Section \ref{sec:sitdemoexp}) for demonstrations containing a single constraint.  The third experiment seeks to evaluate the complete constraint inference method on a situated demonstration containing multiple differing constraints. 

\subsection{Experimental setup}
 We desired to show that our methods can reliably recognize and parameterize constraints through a variety of demonstrator strategies. In all three experiments, data was collected from 9 participants with no prior knowledge of or prior experience with the research recruited from the university student population. The study took 60 minutes and all participants received \$12 as compensation. To acquire the resulting pose and wrench measurements that occur during constraint interactions, subjects used custom instrumented tongs \cite{PSMG19} equipped with force-torque sensors (ATI mini40) and motion capture markers (Opti-track Flex 13) to perform demonstrations.  The instrumented tongs, shown in Fig. \ref{fig:tongs}, provide measured position, orientation, force, and moment trajectories sampled at 120 Hz (after appropriate resampling and filtering). While our method extends to other input methods capable of measuring interaction wrenches, instrumented tongs provides a familiar interface for human demonstrators.
 
 For the first two experiments, where the \emph{model fitting} and \emph{model selection} performance are evaluated, an experimental test bed, incorporating various physical constraints was constructed (see Fig. \ref{fig:testbed}). Each constraint in the test bed has a designated grasp location that participants use to interact with the constraint. 

In all demonstrations except \emph{point-on-plane}, \emph{point on line}, and the situated demonstration, users interact with a non-rigid spherical interface. For constraint models that do not incorporate orientation constraints, such as \emph{prismatic relaxed}, \emph{planar relaxed}, and \emph{axial relaxed} constraint, the measured pose information is taken directly from the instrumented tong motion capture markers.  For constraint models where orientation of the constrained body is enforced, such as \emph{point-on-plane}, \emph{point-on-line}, \emph{prismatic}, \emph{axial}, and \emph{planar} constraints, the pose information is taken from motion capture makers attached directly to the manipulated object. While our method can fit rigid interactions using the pose of the instrumented tongs, users would allow rotation when interfacing with the spherical interface, and using separate markers attached to the object allowed for rigid and relaxed data to be captured from a single demonstration without having to instruct users to perform a separate set of demonstrations where objects were held rigidly. In cases where the object markers are used, the applied forces and moments measured using the instrumented tongs are transformed to the object coordinate frame by relating the object and instrumented tongs motion capture markers.  
\subsection{Evaluation of Model Fitting}\label{sec:fittingexp}
The first experiment seeks to evaluate the performance of constraint \emph{model fitting} (Section \ref{sec:regression}) for demonstrations containing a single constraint. We compare the performance of a method using pose and wrench information to our previous method \cite{Subramani2018,SubramaniRAL2018}, which uses only pose measurements.

Specifically, we evaluate fit accuracy as a function of demonstration duration.  In the experiments, participants were instructed to interact with each physical constraint on the test bed. Two 10 second demonstrations were recorded per constraint.   To simulate varying lengths of constraint interactions, contiguous segments are randomly chosen from across the entire demonstration and fit using the method described in \ref{sec:regression}.
For comparison, the data was also fit to  constraint models using a method that considered pose information only.  In this case, the method is identical to that described in Section \ref{sec:regression} except that the maximum likelihood estimate objective function relies on the kinematic error objective given in equation (\ref{eq:K_error}) only (i.e., does not include the wrench error objective function from equation (\ref{eq:W_error})).

To compare the methods, a fit error based on the position error metric described in equation (\ref{eq:kerrormetricr}) is computed for both methods. While a variety of error metrics are possible, we chose a position-based method to allow comparison across constraint models and as a method that can be related to unexpected applied force through a hybrid controller. The fits are evaluated against either the entire recorded demonstration or for \emph{point-on-line} and \emph{point-on-plane}, an independent demonstration performed by the experimenter (since the fitting requires a significant amount of the original 10 second demonstration).

\begin{figure}[t]
% \includesvg{./figures/regression_figures/regression_planar_relaxed.svg}
% \includesvg{./figures/regression_figures/regression_point_on_plane.svg}
% \includesvg{./figures/regression_figures/regression_point_on_line.svg}
% \includesvg{./figures/regression_figures/regression_planar.svg}
\centering
\includegraphics[width=3.2in]{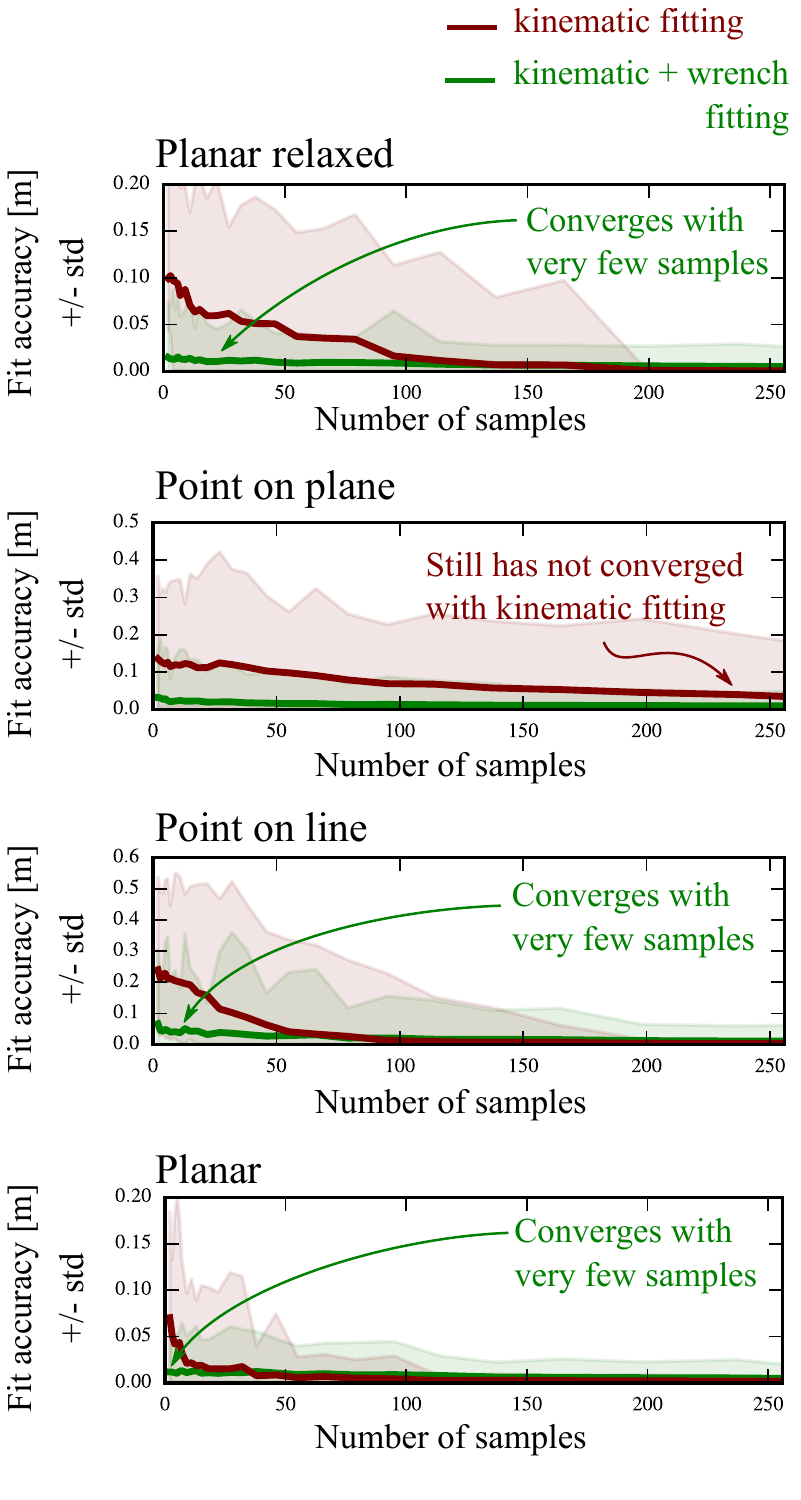}
\caption{Comparison  between  fit  accuracy in meters versus  number  of  contiguous  samples (120 Hz)  in  a  demonstration  for  high degree of freedom models. Fitting with kinematic and wrench measurements [green line]. Fitting with kinematic measurements [red line]. Shaded regions correspond to the first standard deviation about the mean. Constraints with higher degrees of freedom converge with  fewer  samples  when  wrench  measurements  are  incorporated  during  the  fitting  process.}
\label{fig:highdofs}
\end{figure}

Fig. \ref{fig:highdofs} shows a regression analysis for constraints with large number of degrees of freedom.
Fit accuracy is evaluated as a function of contiguous samples (sampled at 120 Hz).
For models with high degrees of freedom (\emph{point-on-plane} - 5 DOF, \emph{planar relaxed} - 4 DOF, \emph{planar} - 3 DOF, and \emph{point-on-line} - 4 DOF) considerably fewer samples of combined kinematic and wrench data are required to fit models than for kinematic (i.e., pose) data alone.

%Frequently \emph{point on plane} constraint does not fit at all when using only kinematic measurements. This is shown by the large standard deviation when incorporating 

\begin{figure}[h]
\centering
\includegraphics[width=3.2in]{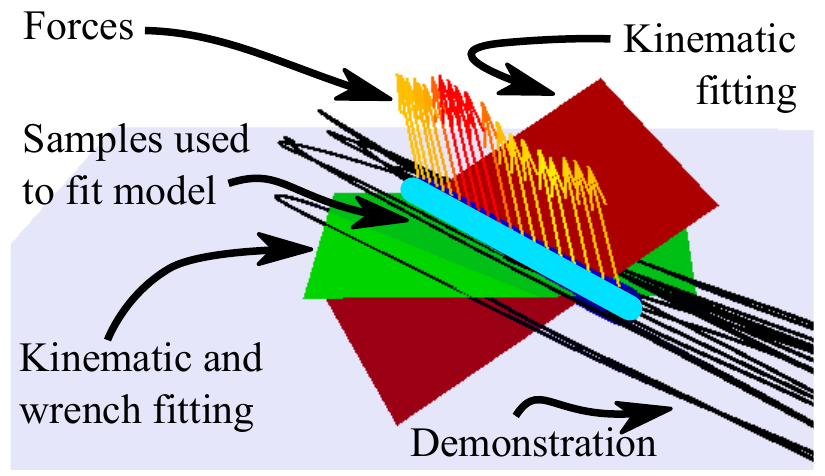}
\caption{An example demonstration of the \emph{planar relaxed constraint} from Demonstrator 6. 
Using a small portion in the demonstration (cyan) of 0.2 seconds fitting with only kinematic data (red plane) does not produce the correct model. Including reaction wrenches in the analysis fits a better plane (green) to the demonstration. }
\label{fig:degeneratePlane}
\end{figure}

Wrench measurements provide additional information that compensates for degenerate kinematic information allowing a constraint to be fit with fewer samples. Fig. \ref{fig:degeneratePlane} shows a comparison between fitting the \emph{planar relaxed constraint} model with pose and wrench measurements compared to fitting with only pose measurements for a particular participant's demonstration. Short demonstrations on the \emph{planar relaxed} constraint may consist of primarily linear motion which leads to degenerate planes when fitting with only kinematic data. Motion in a straight line will not completely determine the orientation of the plane as many different planes may fit the same line.

\begin{figure}[t]
\includegraphics[width=3.2in]{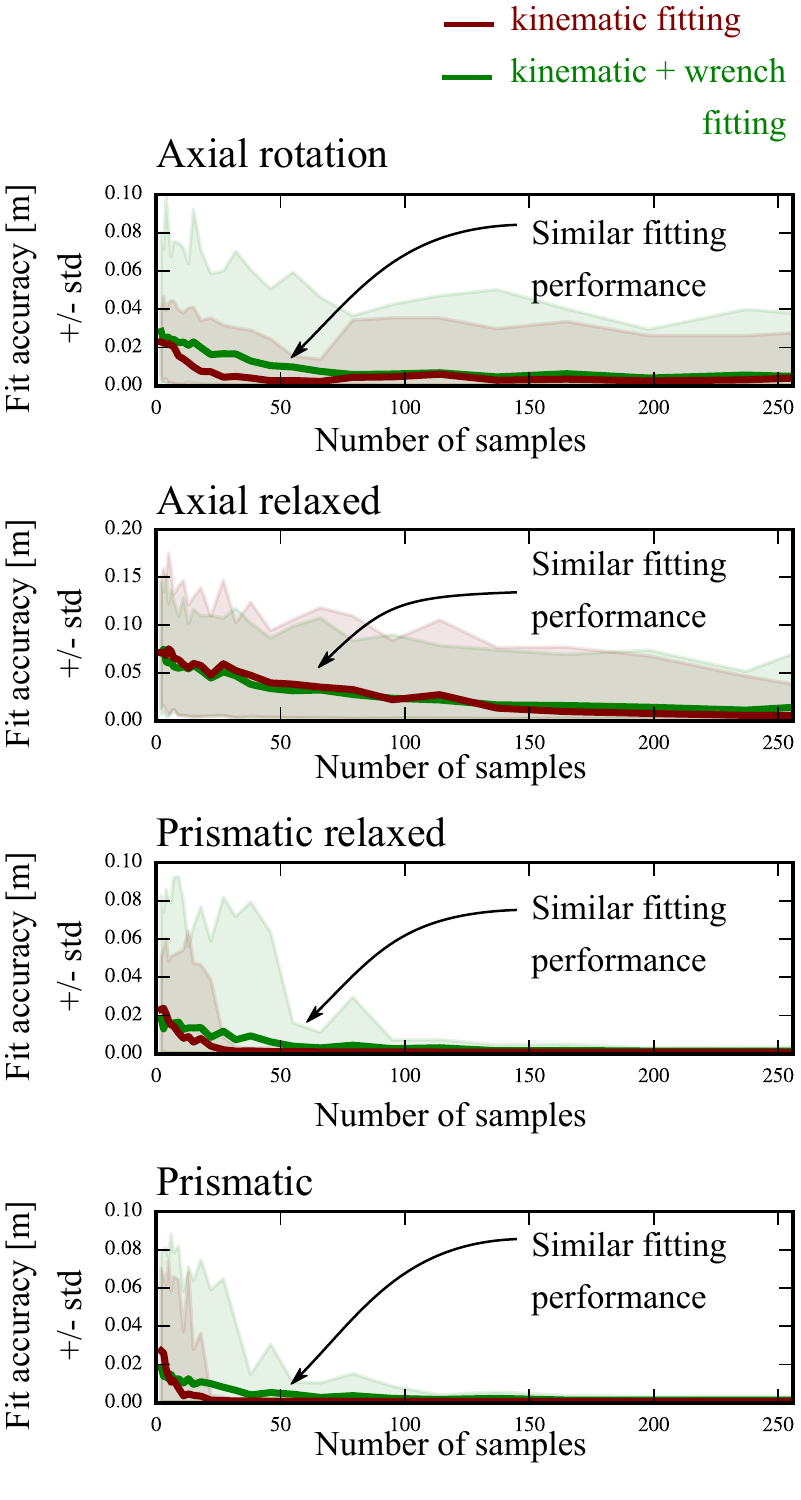}
\caption{Fit  accuracy in meters versus  number  of  contiguous  samples (120 Hz)  in  a  demonstration  for  each  constraint. Fitting with kinematic and wrench measurements [green line]. Fitting with kinematic measurements [red line]. Shaded regions correspond to the first standard deviation about the mean.  Constraints with lower degrees of freedom do not benefit when wrench  measurements  are  incorporated  during  the  fitting  process. }
\label{fig:lowdofs}
\end{figure}

For models with fewer degrees of freedom such as the \emph{prismatic}, \emph{axial rotation}, \emph{axial relaxed}, and \emph{planar} constraints, the wrench method yields similar results to the kinematic-only fitting method. This is expected since high numbers of kinematic constraints lead to many directions of admissible forces and moments which limits the utility in fitting. 
For example, in the \emph{prismatic} model, there is only a single degree of freedom that is not constrained. However, this is the direction of motion which is also the direction in which friction occurs. Since we assume that forces in this direction could be friction, they are removed. The remaining two directions are acceptable reaction forces. As a result, there is no additional information added from including wrench measurements in fitting for this model.
Fig. \ref{fig:lowdofs} shows a regression analysis of number of samples verses fit error for the constraints with a small number of degrees of freedom. 
The \emph{axial relaxed} and \emph{axial rotation} models have high nonlinearity and sometimes the fits are incorrect due to convergence to a local minimum. 
This explains the large standard deviation for both models even when using a large number of samples. 

In this experimental setup, the orientation of the force frame is derived from motion capture orientation measurements. As a result, the force and moment errors tend to be higher based on the propagated orientation error. This can be seen in the increased noise floor for the wrench fitting method in the regression plots (i.e., the wrench method converges to a higher error than the kinematic method). While this is a product of the experimental setup, it demonstrates that it may still be beneficial to consider using kinematic fitting for constraints where the force and moments add little information (e.g., \emph{prismatic constraint}) to avoid cases when the forces and moments have a significant noise floor and to simplify the dimensionality of the non-linear optimization.

The experiments show that wrench measurements improve fitting of models with high degrees of freedom over short demonstration duration. 
They help in situations when kinematic information is insufficient to completely define the constraint model. 
For constraint models with high degrees of freedom, we recommend that wrench measurements are incorporated in the fitting process. For models with low degrees of freedom, wrench measurements can be omitted to decrease computation time and reduce the complexity of the optimization process. 

\subsection{Evaluation of Model Selection}
\label{sec:modelselectionexp}

The second experiment seeks to evaluate the performance of constraint \emph{model selection} (Section \ref{sec:modelselection}) for demonstrations containing a single constraint and compare the performance of a method using both pose and wrench measurements to an approach that uses only pose measurements. For comparison, we also include a penalized kinematic method \cite{akaikeh1974} to address limitations with a kinematic-only method, where demonstration data may be consistent with a number of different kinematic models and thus result in poor selection accuracy. The penalized likelihood metric is defined as follows:
\begin{equation}\label{eq:ms}
    m =  \argmin_{m}(\beta DOF(m) + \gamma \chi -\ln{{L}(m)})
    \end{equation}
where DOF refers to the kinematic degrees of freedom of a model and $\chi$ is the number of model parameters (Corresponding values for the constraint models used in this paper are provided in Appendix \ref{app:models}. Weighted parameters $\beta$ and $\gamma$ are fit using eighty training demonstrations (10 per constraint model). The accuracy is maximized with $\beta=1.69$ and $\gamma=0.40$ (accuracy: 0.65).

\begin{figure*}[ht!]
\centering
\includegraphics[width=7in]{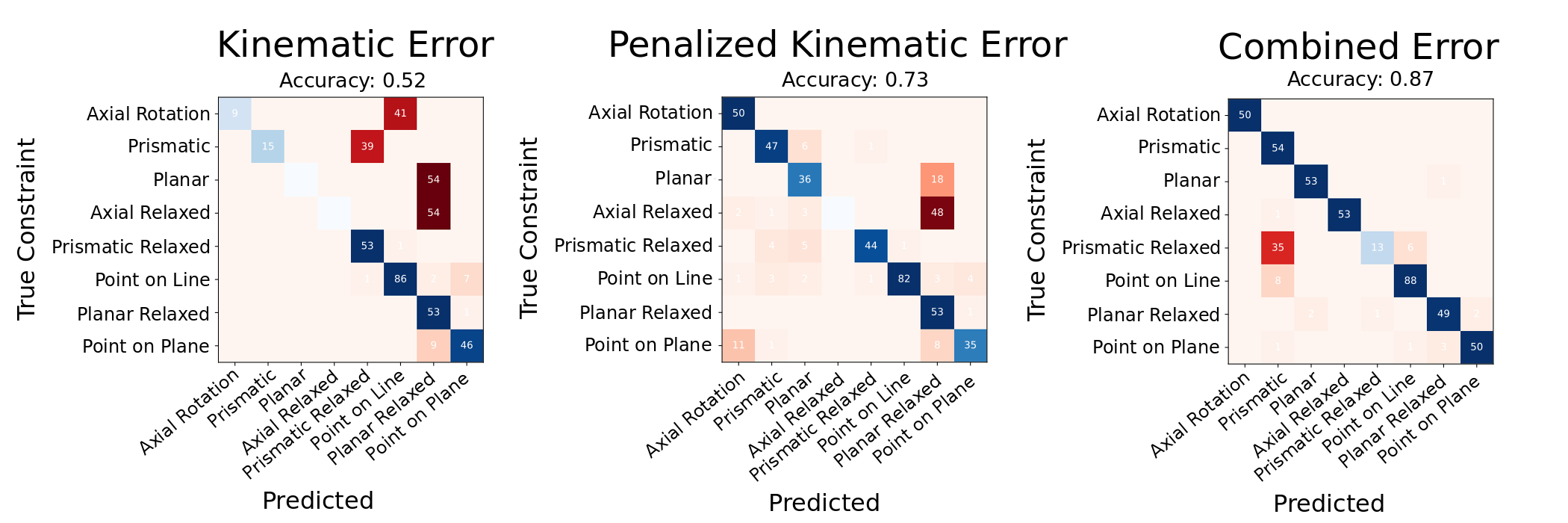}
\caption{Comparison between kinematic and combined methods for model selection. The use of a penalized likelihood greatly improves kinematic model selection, but the method incorporating wrench measurements has a much higher overall classification.}
\label{fig:modelselection}
\end{figure*}

In our experiments, the demonstrations are split into 3.33 second segments (sufficient samples for fitting). Since the demonstrations are sufficiently long, the constraint parameters for all demonstrations are fit with kinematic data to ensure that the differing methods use the same constraint fit parameters. Since model selection is requisite on the constraint being properly fit, fits that do not match the performed constraint are manually discarded. The results appear in Fig. \ref{fig:modelselection}.

The results shown in Fig. \ref{fig:modelselection} were, in part, evaluated using the error tallying method described in equation (\ref{eq:MS}) which, in turn, requires that we set the thresholds for each of the error metrics. The metric values are set based on an estimate of the data collection accuracy. The force and moment error represent the component orthogonal to the permissible force and moment directions. Since the errors are normalized with respect to the measured force and moment, the formulation is equivalent to evaluating the sine of the angle between the measured and normal vectors. In our experimental evaluation, the error for the force and moment is set to the same value as the orientation error (2 degrees). The list of parameter values is provided in Table \ref{table:MSthresh} below.

\begin{table}[H]
\centering
\begin{tabular}{|l|l|}
\hline
\textbf{Threshold}   & \textbf{Value}  \\ \hline
Position    & 2 mm     \\ \hline
Orientation & 2 deg         \\ \hline
Force       & 0.06 (sin(2 deg))      \\ \hline
Moment      & 0.06 (sin(2 deg))      \\ \hline 
\end{tabular}
\caption{Threshold values for error tallying}
\label{table:MSthresh}
\end{table}

 Our results demonstrate the value of including wrenches in constraint model selection. As seen in Figure \ref{fig:modelselection}, the method including wrench information in model selection greatly outperforms equivalent kinematic and penalized-kinematic voting methods. In the kinematic-only tallying method, incorrect classifications can be attributed to model generality. For example, the \emph{axial relaxed} constraint is always identified as \emph{planar relaxed}. When only kinematic measurements are considered, this is a likely case. There is always a plane that can describe an axial motion and moreover any small errors in the position might still be consistent with the plane model, but not necessarily with the axial model. Two other models, (\emph{planar constraint} and \emph{prismatic constraint}), are consistently misidentified as their relaxed versions. As a reminder, the relaxed version of the constraint does not enforce a rigid orientation. Similar to the previous case, any small errors when the orientation is rigid favor the relaxed model in the selection process. The kinematic method also misidentifies the rigid axial constraint as a point on line constraint. In this case, both models do not receive any tallies and as a result, the selection is arbitrary based on the order in which the models are selected. Penalization of the kinematic method increases classification accuracy from 0.52 to 0.73 correcting for many of the issues with the general model. However, even with correction, the \emph{axial relaxed} model is still identified as a \emph{planar relaxed} constraint.

Use of the combined error greatly improves the classification accuracy (0.87). The only model that is consistently misidentified is the \emph{prismatic relaxed constraint}. In most cases, the identified model is the rigid version of the constraint. The relaxed model does not enforce an orientation which consequently means there can be no generated moments. When a participant interacts with the tongs, the grip can slip leading to changes in orientation, but for the most part, the generated moments and rigidity may be more consistent with the rigid model. As such, these cases are not necessarily incorrect classifications but instead  represent cases of demonstrations that lie between the two constraint models.
\subsection{Situated Demonstration}\label{sec:sitdemoexp}
We evaluate our system on a proof-of-concept situated task. The task involves preparing an espresso using the instrumented tongs as an input device. The experimental setup for the espresso preparation task in shown in Fig. \ref{fig:teaserEspressoTask}.

\begin{enumerate}
\item The demonstrator first slides an espresso cup under the espresso machine spout (\emph{planar-relaxed constraint}). 
\item The demonstrator then uses the instrumented tongs to pull out a drawer (\emph{prismatic-relaxed constraint}) containing a single coffee pod. 
\item Using the tongs the demonstrator carries the pod to the espresso machine and places the pod inside. 
\item Finally the demonstrator uses the tongs to close the lever (\emph{axial-relaxed constraint}) of the espresso machine.  
\end{enumerate}

Using the system described in this paper, the demonstration is parsed for the three possible constraint models: \emph{planar relaxed}, \emph{prismatic relaxed}, and \emph{axial relaxed} constraints. The three relaxed models are used as the instrumented tongs have slip directions and do not enforce a rigid orientation when grasping the constrained objects. The two 1-DOF models, \emph{prismatic relaxed} and \emph{axial relaxed} models, are fit using the kinematic method, to avoid the increased noise floor of the experimental setup.

\begin{figure}[h]
\centering
\includegraphics[width=3in]{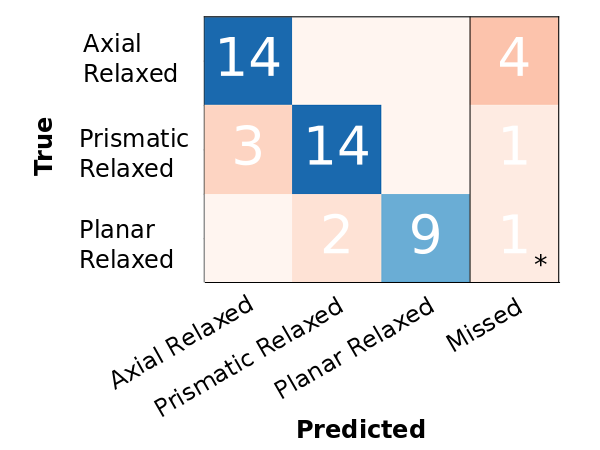}
\caption{
Confusion matrix of the end-to-end system on 18 espresso making demonstrations (9 participants, 2 trials, 3 potential constraints per task). In 6 demonstrations, the participant lifted the cup rather than sliding it, so there was no planar constraint to be detected (*).
}
\label{fig:espressoconfusion}
\end{figure}

The results of the constraint inference are shown in Fig. \ref{fig:espressoconfusion}. The system correctly identifies 37 out of the 48 constraints in 18 demonstrations (2 per participant). 
%% MG explain 58 constraints
Note, data was removed in instances where participants did not correctly interact with a given constraint (e.g. participants lifted the cup rather than sliding it on the table).

The espresso setup evaluates the method on a realistic task where the constraints may not be as precise as the constraints in the earlier experiments (using the experimental test bed shown in Fig. \ref{fig:testbed}).
For example, the drawer used for linear motion is plastic and has 1.5 cm of play in its constrained motion. Additionally, the espresso machine lever (\emph{axial relaxed constraint}) can be difficult to grasp leading to some demonstrations that do not resemble axial motion.
% The constraints from the espresso setup are not as clean as those collected from the constraint board. 
As expected, a few constraint interactions were misclassified by model selection or rejected by the null hypothesis.
In spite of the imprecise constraints, the system correctly infers a majority of the constraints.

The null hypothesis was successful in rejecting non-constrained motion that exceeded the force threshold, such as when participants hold the coffee pod and accelerate towards the espresso maker. In the 18 demonstrations, the null hypothesis rejected 20 extraneous segments.

Fig. \ref{fig:teaserEspressoTask} shows the results for one of the participants where the method is successful. 
Our approach correctly selected the model 77\% of the time. Correctly inferred constraints were all correctly fit (as shown in Fig \ref{fig:teaserEspressoTask}). 
Constraints were identified using a single demonstration in which the demonstrator performed them naturally with minimal instruction. 

\section{Conclusion}
The limitations of our method suggest future areas of extension. In our methods, we assume free space interaction occurs between each of the constrained segments. While we have not explored how often this assumption is valid, other work has developed methods for detecting changes between consecutive constraints.

In this work, we focus on methods for constraint inference, however, we do not explore task replication by a robot. While our method provides a general set of tools to recognize constraints and their associated parameters, it is unanswered whether these models appropriately define a task such that a robot could perform similar actions.

Additionally, this work does not discuss human procedures that may be identified and leveraged during replay. We also do not explore remapping of constrained motions between a general input method and a robot which may have different kinematic restrictions. Our method describes when constraints occur and their parameters for a given demonstration. Future work will investigate editing these representations, for example to modify a motion on a constraint.

In summary, we present a framework that is capable of recognizing geometric constraints in a demonstration. Specifically, the method includes the ability to segment sections of constrained motion, fit models using both kinematic and wrench measurements, and select the best candidate model for each section. We show that wrench measurements enable estimation of constraints with high degrees of freedom from fewer samples compared to using only kinematic measurements. We demonstrate the use of wrench measurements in improving model selection classification accuracy compared to a similar method using only kinematic measurements. We demonstrate that model selection can be further improved by incorporating model generality through the degrees of freedom and number of parameters of candidate constraint models. We evaluate our end-to-end constraint inference method on a situated task of preparing an espresso which correctly identifies 77\% of constraints without any user input.

\appendices
\section{Mathematical models of geometric constraints}\label{app:models}
This section describes the mathematical model development for the various constraint types evaluated in the paper (see Fig. \ref{fig:constraints}). The mathematical model for the \emph{point-on-plane constraint} is provided in section \ref{sec:point_on_plane_model}.

For all constraint types, our mathematical model considers a 6-degree of freedom rigid body of negligible inertial properties located in space through 3 translational coordinates $\mathbf{r} \in \mathbb{R}^3$  and rotational coordinates represented either by a unit quaternion $\mathbf{q} \in$   Spin(3) or an orthogonal rotational matrix $\mathbf{A}(\mathbf{q}) \in \text{SO(3)}$. Together $(\mathbf{r},\mathbf{q})$ form the body coordinates $\mathbf{p}$. Each constraint type has a set of parameters, $\boldsymbol{\alpha}$, that parameterize the geometry of the constraint.

% Let the parameters of the constraint be  then equation $\Phi$ becomes $\Phi(\mathbf{p},\mathbf{\alpha})$.

\subsection{Planar constraint} The planar constraint is exemplified by an eraser moving against a whiteboard. The rigid body interacting with the plane (e.g. eraser) can only rotate about a vector perpendicular to the plane, and all points within the rigid body translate parallel to this plane. 

A plane may be specified by applying a general displacement (i.e. translation and rotation) transformation of the \emph{x-y} plane which involves: 
\begin{enumerate}
\item translating the \emph{x-y} coordinate plane along the \emph{z}-axis
\item rotating the translated plane about the origin. 
\end{enumerate}
We represent the rotation transformation using two exponential coordinates $\textbf{w}=[{{w}_{x}},{{w}_{y}},0]\in \mathbb{R}^2$ corresponding to ${{e}^{{\tilde{\textbf{w}}}}}\in SO(3)$, the exponential map, which is equivalent to a rotation matrix with an axis of rotation in the \emph{x-y} plane. Rodrigues' rotation formula \cite{murray1994mathematical} is used to compute ${e}^{\tilde{\mathbf{w}}}$. $\tilde{\mathbf{w}}$ is the skew symmetric matrix similar to equation (\ref{eq:skewsym}) in Appendix \ref{app:ki}. The third term of \textbf{w} is zero because rotations about the \emph{z} - axis (perpendicular to the plane) do not alter the plane's geometry. The translation is represented by $d \in \mathbb{R}$. The normal vector on this plane is represented by ${e}^{\tilde{\mathbf{w}}}{\left[ 0\ \ 0\ \ 1 \right]}^{T}$ and the shifted origin of the \emph{x-y} plane is represented by ${{e}^{{\tilde{\mathbf{w}}}}}{{\left[ 0\ \ 0\ \ d \right]}^{T}}$.

The origin of the local coordinate frame on the body is coincident with the plane.  The constraint equations for motion on a plane is given as:
\begin{equation}\label{eq:planarrelaxed}
{{\Phi }_{1}}\equiv {{\left( {{e}^{{\tilde{\mathbf{w}}}}}{{\left[ 0\ \ 0\ \ d \right]}^{T}}-\mathbf{r} \right)}^{T}}{{e}^{{\tilde{\mathbf{w}}}}}{{\left[ 0\ \ 0\ \ 1 \right]}^{T}}=0
\end{equation}

\noindent A unit vector $\bar{\mathbf{t}}$ is defined to force the rigid body perpendicular to the plane: 
\begin{gather}
{{\Phi }_{2}}\equiv {{\left( \textbf{A}(\mathbf{q})\bar{\mathbf{t}}~ \right)}^{T}}{{e}^{{\tilde{\mathbf{w}}}}}{{\left[ 1\ \ 0\ \ 0 \right]}^{T}}=0\label{eq:p1}\\
{{\Phi }_{3}}\equiv {{\left( \textbf{A}(\mathbf{q})\bar{\mathbf{t}}~ \right)}^{T}}{{e}^{{\tilde{\mathbf{w}}}}}{{\left[ 0\ \ 1\ \ 0 \right]}^{T}}=0\label{eq:p2}\\
{{\Phi }_{\mathbf{p}1}}\equiv {{\bar{\mathbf{t}}}^{T}}\bar{\mathbf{t}}-1=0 \label{eq:c6}
\end{gather}

\noindent Unity of $\bar{\mathbf{t}}$ is enforced by equation (\ref{eq:c6}). The parameters of this constraint are $\boldsymbol{\alpha}  =  (\bar{\mathbf{t}},d,{{w}_{x}},{{w}_{y}})$. The rigid body has 3 degrees of freedom.

\noindent\textbf{Friction Model:} The friction force is in the direction of linear velocity and the friction moment is in the direction of angular velocity. 

\begin{gather}
    \mathbf{F}_{\mu} = (\mathbf{F} \cdot\widehat{\mathbf{v}})\widehat{\mathbf{v}} \label{eq:frictionforceplanar}\\
    \mathbf{N}_{\mu} = (\mathbf{N} \cdot\widehat{\boldsymbol{\omega}})\widehat{\boldsymbol{\omega}} \label{eq:frictionmomenteplanar}
\end{gather}

where $\mathbf{F}$ and $\mathbf{N}$ are the measured force and moment, respectively. The measured linear and angular velocity are $\mathbf{v}$ and $\boldsymbol{\omega}$, respectively.

\subsection{Planar relaxed constraint}
The \emph{planar relaxed constraint} is represented by equation (\ref{eq:planarrelaxed}). The friction model is the same as the \emph{planar constraint}.

\subsection{Prismatic constraint}
The prismatic constraint represents translational motion in one direction. It is similar to pulling out a drawer. All points on the rigid body translate identically. 
This constraint requires a representation of the axis of translation. Similar to the plane, the axis of translation can be generated by applying a general displacement (i.e. translation and rotation) transformation of the \emph{z} coordinate axis which is equivalent to:
\begin{enumerate}
\item translating the \emph{z} axis in the \emph{x-y} plane
\item rotating the translated axis about the origin. 
\end{enumerate}

This is represented by two exponential coordinates  $\textbf{w}=[{{w}_{x}},{{w}_{y}},0]\in \mathbb{R}^2$ and two translational coordinates $d_x , d_y \in \mathbb{R}$. The third term in \textbf{w} is zero because rotations about the \emph{z} - axis produce a line that could be produced by an alternative translation motion. This defines the axis in the global reference frame. The tangent to this axis is ${{e}^{{\tilde{\mathbf{w}}}}}{{\left[ 0\ \ 0\ \ 1 \right]}^{T}}$ and the translated origin is ${{e}^{{\tilde{\mathbf{w}}}}}{{\left[ {{d}_{x}}~{{d}_{y}}~0 \right]}^{T}}$. ${{e}^{{\tilde{\mathbf{w}}}}}{{\left[ 1\ \ 0\ \ 0 \right]}^{T}}$ and ${{e}^{{\tilde{\mathbf{w}}}}}{{\left[ 0\ \ 1\ \ 0 \right]}^{T}}$  represent vectors perpendicular to this axis. The origin of the local coordinate frame of the body is coincident with this line/axis:
\begin{gather}
{{\Phi }_{1}}\equiv {{\left( {{e}^{{\tilde{\mathbf{w}}}}}{{\left[ {{d}_{x}}~{{d}_{y}}~0 \right]}^{T}}-\mathbf{r} \right)}^{T}}{{e}^{{\tilde{\mathbf{w}}}}}{{\left[ 1\ \ 0\ \ 0 \right]}^{T}}=0\label{eq:l1}\\
{{\Phi }_{2}}\equiv {{\left( {{e}^{{\tilde{\mathbf{w}}}}}{{\left[ {{d}_{x}}~{{d}_{y}}~0 \right]}^{T}}-\mathbf{r} \right)}^{T}}{{e}^{{\tilde{\mathbf{w}}}}}{{\left[ 0\ \ 1\ \ 0 \right]}^{T}}=0\label{eq:l2}
\end{gather}

\noindent The orientation of the rigid body is fixed using the following equations: 
\begin{gather}
{{\Phi }_{3}}\equiv {{(\textbf{A}(\mathbf{q})\bar{\mathbf{s}})}^{T}}{{e}^{{\tilde{\mathbf{w}}}}}{{\left[ 1\ \ 0\ \ 0 \right]}^{T}}=0\label{eq:l3}\\
{{\Phi }_{4}}\equiv {{(\textbf{A}(\mathbf{q})\bar{\mathbf{s}})}^{T}}{{e}^{{\tilde{\mathbf{w}}}}}{{\left[ 0\ \ 1\ \ 0 \right]}^{T}}=0\label{eq:l4}\\
{{\Phi }_{5}}\equiv {{(\textbf{A}(\mathbf{q})\bar{\mathbf{t}})}^{T}}{{e}^{{\tilde{\mathbf{w}}}}}{{\left[ 1\ \ 0\ \ 0 \right]}^{T}}=0\label{eq:l5}\\
{{\Phi }_{\mathbf{p}1}}\equiv {{\bar{\mathbf{s}}}^{T}}\bar{\mathbf{s}}-1=0\label{eq:l6}\\
{{\Phi }_{\mathbf{p}2}}\equiv {{\bar{\mathbf{s}}}^{T}}\bar{\mathbf{t}}=0\label{eq:c5}
\end{gather}
\noindent where $\bar{\mathbf{s}}$  and $\bar{\mathbf{t}}$  are unit vectors fixed on the rigid body. Equations (\ref{eq:l3}) through (\ref{eq:c6}) and (\ref{eq:l6}) prevent the body from rotating. Equations (\ref{eq:l6}) and (\ref{eq:c6}) forces unity of $\mathbf{s}$ and $\mathbf{t}$ respectively. 
This rigid body has one degree of freedom of motion. The constraint parameters are: $\boldsymbol{\alpha}  =  (\bar{\mathbf{t}},\bar{\mathbf{s}},{{d}_{x}},{{d}_{y}},{{w}_{x}},{{w}_{y}})$.

\noindent\textbf{Friction Model:} Friction force is in the direction of linear velocity. This model does not permit angular rotation so we assume zero friction moment.  

\begin{gather}
    \mathbf{F}_{\mu} = (\mathbf{F} \cdot\widehat{\mathbf{v}})\widehat{\mathbf{v}} \label{eq:frictionforceprismatic}\\
    \mathbf{N}_{\mu} = 0 \label{eq:frictionmomenteprismatic}
\end{gather}

where $\mathbf{F}$ is the measured force and $\mathbf{v}$ is the linear velocity.
\subsection{Prismatic relaxed constraint}
The \emph{prismatic relaxed constraint} is obtained by only using equations (\ref{eq:l1}) and (\ref{eq:l2}). The friction model is given by equations (\ref{eq:frictionforceprismatic}) and (\ref{eq:frictionmomenteplanar}).

\subsection{Point-on-line constraint}
The \emph{point-on-line constraint} represents a point on a rigid body constrained to a line in space. It is similar to a pen moving against a straight edge.

Consider a point $\mathbf{s^*} \in \mathbb{R}^3$ in the global reference frame defined as a fixed point on the rigid body. $\mathbf{s}$ is a vector directed from the origin of the local reference frame of the body to the rigidly fixed point $\mathbf{s^*}$ in the global reference frame. The local reference frame counterpart of $\mathbf{s}$ is $\overset{-}{\mathbf{s}}$ :
\begin{equation}
\mathbf{s*}\equiv \mathbf{r}+\mathbf{s}\equiv \mathbf{r}+\textbf{A}(\mathbf{q})\overset{-}{\mathbf{s}}
\end{equation}
\noindent where \textbf{A}(q) is the orthogonal rotation matrix equivalent to the rotation represented by q.
$\mathbf{s^*}$ on the rigid body is coincident with a line represented by parameters (${{d}_{x}},{{d}_{y}},{{w}_{x}},{{w}_{y}}$):

\begin{gather}
{{\Phi }_{1}}\equiv {{\left( {{e}^{{\tilde{\mathbf{w}}}}}{{\left[ {{d}_{x}}~{{d}_{y}}~0 \right]}^{T}}-\mathbf{r} - \textbf{A}(\mathbf{q})\overset{-}{\mathbf{s}} \right)}^{T}}{{e}^{{\tilde{\mathbf{w}}}}}{{\left[ 1\ \ 0\ \ 0 \right]}^{T}}=0\label{eq:lp1}\\
{{\Phi }_{2}}\equiv {{\left( {{e}^{{\tilde{\mathbf{w}}}}}{{\left[ {{d}_{x}}~{{d}_{y}}~0 \right]}^{T}}-\mathbf{r} - \textbf{A}(\mathbf{q})\overset{-}{\mathbf{s}} \right)}^{T}}{{e}^{{\tilde{\mathbf{w}}}}}{{\left[ 0\ \ 1\ \ 0 \right]}^{T}}=0\label{eq:lp2}
\end{gather}
\\
\textbf{Friction Model:} The friction models are identical to the 
friction models for the \emph{point-on-plane constraint} described in section \ref{sec:point_on_plane_model}. 

\subsection{Axial rotation constraint} The axial rotation constraint is similar to a door knob or a hinged door, all points on the rigid body rotate about an axis and translations are not permitted.

Consider a point on the rigid body $\mathbf{s^*}$ that is on the axis of rotation and in the plane perpendicular to the axis containing the coordinate frame origin. A rigid point on the axis of rotation defined as $({{d}_{x}},{{d}_{y}},{{d}_{z}})\in {\mathbb{R}^{3}}$ constrains the point $\mathbf{s^*}$. The vector  $\bar{\mathbf{s}}$ is perpendicular to this axis. The rigid body must not rotate about vector $\bar{\mathbf{s}}$ so unit vector $\bar{\mathbf{t}}$ is introduced to enforce this condition. The constraint equations are (\ref{eq:c6}), (\ref{eq:c5}) and the following:
\begin{gather}
{{\Phi }_{1}}\equiv ~\mathbf{r}+\textbf{A}(\mathbf{q})\bar{\mathbf{s}}-{{[{{d}_{x}},{{d}_{y}},{{d}_{z}}]}^{T}}=0\label{eq:ax1}\\
{{\Phi }_{2}}\equiv {{\left( \textbf{A}(\mathbf{q})\bar{\mathbf{t}}~ \right)}^{T}}{{e}^{{\tilde{\mathbf{w}}}}}{{\left[ 0\ \ 0\ \ 1 \right]}^{T}}=0 \label{eq:ax2}\\
{{\Phi }_{3}}\equiv {{\left( \textbf{A}(\mathbf{q})\bar{\mathbf{s}}~ \right)}^{T}}{{e}^{{\tilde{\mathbf{w}}}}}{{\left[ 0\ \ 0\ \ 1 \right]}^{T}}=0 \label{eq:ax3}
\end{gather}

Constraint parameters are $\boldsymbol{\alpha}  =  (\bar{\mathbf{t}},\bar{\mathbf{s}},{{d}_{x}},{{d}_{y}},{{d}_{z}},{{w}_{x}},{{w}_{y}})$. This constraint has 1 degree of freedom.

\noindent\textbf{Friction Model:} The friction wrenches and reaction wrenches are coupled in this model. The principle of virtual work enables the computation of reaction forces and reaction wrenches from the corresponding measured quantities. Let's assume all measured forces are reaction forces. This implies that friction forces are zero: 
\begin{equation}
\mathbf{F}_{\mu} = 0 \label{eq:frictionforceaxial} 
\end{equation}
\noindent The principle of virtual work requires that wrenches are orthogonal to twists as per Equation \ref{eq:vw} (substituting virtual displacements with velocities). The residual of equation \ref{eq:vw} provides the corresponding friction work $W_\mu$: 
\begin{equation}
W_\mu = \mathbf{v} \cdot \mathbf{F} + \boldsymbol{\omega} \cdot \mathbf{N} \label{eq:frictionmomentaxialwork} 
\end{equation}
\noindent The friction work comes from the friction moment about the axis of rotation. The friction moment is directed about the axis of rotation. 

\begin{equation}
\mathbf{N}_\mu = \frac{W_\mu}{|\boldsymbol{\omega}|}\hat{\boldsymbol{\omega}} \label{eq:frictionmomentaxial} 
\end{equation}

\subsection{Axial relaxed constraint}
The \emph {axial relaxed constraint} also has parameters (${{d}_{x}},{{d}_{y}},{{d}_{z}},{{w}_{x}},{{w}_{y}}$) to configure the axis of rotation and the center of rotation. The radius $l \in {\mathbb{R}}$ is included to fully define the constraint.
\begin{gather}
{{\Phi }_{2}}\equiv  \left\lVert \mathbf{r} -{{[{{d}_{x}},{{d}_{y}},{{d}_{z}}]}^{T}} \right\lVert - \mathbf{l}^2 =0 \label{eq:axrelaxed1}\\
{{\Phi }_{1}}\equiv \left( \mathbf{r} -{{[{{d}_{x}},{{d}_{y}},{{d}_{z}}]}^{T}}\right)^T{{e}^{{\tilde{\mathbf{w}}}}}{{\left[ 0\ \ 0\ \ 1 \right]}^{T}} =0\label{eq:axrelaxed2}
\end{gather}
Equation (\ref{eq:axrelaxed1}) restricts the body origin to a sphere of radius $l$ centered about $(d_x,d_y,d_z)$. 
Equation (\ref{eq:axrelaxed2}) restricts the body origin to a plane containing $(d_x,d_y,d_z)$ with normal ${{e}^{{\tilde{\mathbf{w}}}}}{{\left[ 0\ \ 0\ \ 1 \right]}^{T}}$. 

The friction model is represented by equations (\ref{eq:frictionforceplanar}) and (\ref{eq:frictionmomenteplanar})

\subsection{Degrees of freedom and model parameters per constraint}
Table \ref{table:Modelparams} shows the degrees of freedom and number of constraint parameters for each of the models considered in this paper. These values are used to construct the penalized likelihood method used for comparison in section \ref{sec:modelselectionexp}.
\begin{table}[H]
\centering
\begin{tabular}{|l|c|c|}
\hline
\textbf{Constraint}   & \textbf{DOF} & \textbf{number of model parameters}  \\ \hline
Point-on-plane    & 5 & 6     \\ \hline
Point-on-line & 4 & 7         \\ \hline
Planar       & 3 & 6      \\ \hline
Planar relaxed & 5 & 3    \\ \hline
Prismatic      & 1 & 10      \\ \hline
Prismatic relaxed & 4 & 4    \\ \hline
Axial rotation & 1 & 11    \\ \hline
Axial relaxed & 4 & 8 \\ \hline

\end{tabular}
\caption{Degrees of freedom (DOF) and number of model parameters for each constraint}
\label{table:Modelparams}
\end{table}
\section{Relationship between $\pi$ and q}\label{app:ki}
Evaluating the partial derivatives of the constraint equations, $\Phi$, is required to determine admissible reaction forces and moments for a given constraint. The partial derivative with respect to rotation uses the virtual rotation variable, $\boldsymbol{\pi}$. This section describes the derivation of $\boldsymbol{\pi}$ in terms of the known orientation, $\mathbf{q}$.

\begin{equation}
{{\Phi}_{\pi }}=~{{\Phi}_{\text{q}}}\frac{1}{2}{{\textbf{G}}^{T}}{{\textbf{A}}^{T}}
\end{equation}
where the rotation is described by quaternion $\mathbf{q}=\left[ {{e}_{0}}~~{{\text{e}}^{T}} \right]^T$ and where ${{e}_{0}}={{q}_{0}}$ and $\mathbf{e}={{\left[ {{q}_{1}}~{{q}_{2}}~{{q}_{3}} \right]}^{T}}$. $\mathbf{A}$ is the orthogonal rotation matrix and $\mathbf{G}$ is a matrix constant.  
\begin{equation}
\textbf{A}=:\left( e_{0}^{2}-{{\mathbf{e}}^{T}}\mathbf{e} \right)I+2\mathbf{e}{{\mathbf{e}}^{T}}+2{{e}_{0}}\tilde{\mathbf{e}}
\end{equation}
\begin{equation}
\textbf{G}=:~\left[ \begin{matrix}
-\mathbf{e} & -\tilde{\text{e}}+{{e}_{0}}\textbf{I}  \\
\end{matrix} \right]
\end{equation}
where 
\begin{equation}
\tilde{\mathbf{e}}=\left[ \begin{matrix}
0 & -{{e}_{z}} & {{e}_{y}}  \\
{{e}_{z}} & 0 & -{{e}_{x}}  \\
-{{e}_{y}} & {{e}_{x}} & 0  \\
\end{matrix} \right]
\label{eq:skewsym}
\end{equation}

and \textbf{I} is the identity matrix
\section*{Acknowledgment}
This work was
supported in part by NSF award 1830242 and the University of Wisconsin-
Madison Office of the Vice Chancellor for Research and Graduate Education
with funding from the Wisconsin Alumni Research Foundation.

% Can use something like this to put references on a page
% by themselves when using endfloat and the captionsoff option.
\ifCLASSOPTIONcaptionsoff
  \newpage
\fi

% trigger a \newpage just before the given reference
% number - used to balance the columns on the last page
% adjust value as needed - may need to be readjusted if
% the document is modified later
%\IEEEtriggeratref{8}
% The "triggered" command can be changed if desired:
%\IEEEtriggercmd{\enlargethispage{-5in}}

% references section

% can use a bibliography generated by BibTeX as a .bbl file
% BibTeX documentation can be easily obtained at:
% http://mirror.ctan.org/biblio/bibtex/contrib/doc/
% The IEEEtran BibTeX style support page is at:
% http://www.michaelshell.org/tex/ieeetran/bibtex/
%\bibliographystyle{IEEEtran}
% argument is your BibTeX string definitions and bibliography database(s)
%\bibliography{IEEEabrv,../bib/paper}
%
% <OR> manually copy in the resultant .bbl file
% set second argument of \begin to the number of references
% (used to reserve space for the reference number labels box)

\bibliography{references.bib}

% Generated by IEEEtran.bst, version: 1.14 (2015/08/26)
\begin{thebibliography}{10}
\providecommand{\url}[1]{#1}
\csname url@samestyle\endcsname
\providecommand{\newblock}{\relax}
\providecommand{\bibinfo}[2]{#2}
\providecommand{\BIBentrySTDinterwordspacing}{\spaceskip=0pt\relax}
\providecommand{\BIBentryALTinterwordstretchfactor}{4}
\providecommand{\BIBentryALTinterwordspacing}{\spaceskip=\fontdimen2\font plus
\BIBentryALTinterwordstretchfactor\fontdimen3\font minus
  \fontdimen4\font\relax}
\providecommand{\BIBforeignlanguage}[2]{{%
\expandafter\ifx\csname l@#1\endcsname\relax
\typeout{** WARNING: IEEEtran.bst: No hyphenation pattern has been}%
\typeout{** loaded for the language `#1'. Using the pattern for}%
\typeout{** the default language instead.}%
\else
\language=\csname l@#1\endcsname
\fi
#2}}
\providecommand{\BIBdecl}{\relax}
\BIBdecl

\bibitem{Berenson2009}
\BIBentryALTinterwordspacing
D.~Berenson, S.~S. Srinivasa, D.~Ferguson, and J.~J. Kuffner, ``{Manipulation
  planning on constraint manifolds},'' \emph{2009 IEEE International Conference
  on Robotics and Automation}, vol.~i, pp. 625--632, 2009. [Online]. Available:
  \url{https://ieeexplore.ieee.org/document/5152399}
\BIBentrySTDinterwordspacing

\bibitem{Stilman2007}
M.~Stilman, ``{Task constrained motion planning in robot joint space},''
  \emph{IEEE International Conference on Intelligent Robots and Systems}, pp.
  3074--3081, 2007.

\bibitem{Huh2018}
\BIBentryALTinterwordspacing
J.~Huh, B.~Lee, and D.~D. Lee, ``{Constrained Sampling-Based Planning for
  Grasping and Manipulation},'' in \emph{2018 IEEE International Conference on
  Robotics and Automation (ICRA)}, 5 2018, pp. 223--230. [Online]. Available:
  \url{https://ieeexplore.ieee.org/abstract/document/8461265}
\BIBentrySTDinterwordspacing

\bibitem{Raibert1981}
M.~H. Raibert and J.~J. Craig, ``{Hybrid Position/Force Control of
  Manipulators},'' \emph{Journal of Dynamic Systems, Measurement, and Control},
  vol. 103, no.~2, p. 126, 1981.

\bibitem{Argall2009}
\BIBentryALTinterwordspacing
B.~D. Argall, S.~Chernova, M.~Veloso, and B.~Browning, ``{A survey of robot
  learning from demonstration},'' \emph{Robotics and Autonomous Systems},
  vol.~57, no.~5, pp. 469--483, 2009. [Online]. Available:
  \url{https://doi.org/10.1016/j.robot.2008.10.024}
\BIBentrySTDinterwordspacing

\bibitem{PSMG19}
\BIBentryALTinterwordspacing
P.~Praveena, G.~Subramani, B.~Mutlu, and M.~Gleicher, ``{Characterizing Input
  Methods for Human-to-robot Demonstrations},'' in \emph{2019 14th ACM/IEEE
  International Conference on Human-Robot Interaction (HRI)}.\hskip 1em plus
  0.5em minus 0.4em\relax IEEE, 3 2019, pp. 344--353. [Online]. Available:
  \url{http://graphics.cs.wisc.edu/Papers/2019/PSMG19}
\BIBentrySTDinterwordspacing

\bibitem{Leopoldo2018}
\BIBentryALTinterwordspacing
L.~Armesto, J.~Moura, V.~Ivan, M.~S. Erden, A.~Sala, and S.~Vijayakumar,
  ``{Constraint-aware learning of policies by demonstration},'' \emph{The
  International Journal of Robotics Research}, vol.~37, no. 13-14, pp.
  1673--1689, 2018. [Online]. Available:
  \url{https://doi.org/10.1177/0278364918784354}
\BIBentrySTDinterwordspacing

\bibitem{Li2017}
M.~Li, K.~Tahara, and A.~Billard, ``{Learning task manifolds for constrained
  object manipulation},'' \emph{Autonomous Robots}, vol.~42, no.~1, pp. 1--16,
  2017.

\bibitem{arodriguez2008}
A.~Rodr{\'{I}}guez, L.~Basa{\~{N}}ez, and E.~Celaya, ``{A Relational
  Positioning Methodology for Robot Task Specification and Execution},''
  \emph{IEEE Transactions on Robotics}, vol.~24, no.~3, pp. 600--611, 6 2008.

\bibitem{Dutre1996ContactEnergy}
S.~Dutre, H.~Bruyninckx, and J.~De~Schutter, ``{Contact identification and
  monitoring based on energy},'' in \emph{Proceedings - IEEE International
  Conference on Robotics and Automation}, vol.~2.\hskip 1em plus 0.5em minus
  0.4em\relax IEEE, 1996, pp. 1333--1338.

\bibitem{Joris1999}
\BIBentryALTinterwordspacing
J.~de~Schutter, H.~Bruyninckx, S.~Dutr{\'{e}}, J.~de~Geeter, J.~Katupitiya,
  S.~Demey, and T.~Lefebvre, ``{Estimating First-Order Geometric Parameters and
  Monitoring Contact Transitions during Force-Controlled Compliant Motion},''
  \emph{The International Journal of Robotics Research}, vol.~18, no.~12, pp.
  1161--1184, 1999. [Online]. Available:
  \url{https://doi.org/10.1177/02783649922067780}
\BIBentrySTDinterwordspacing

\bibitem{Lefebvre2003PolyhedralMotion}
T.~Lefebvre, H.~Bruyninckx, and J.~Schutter, ``{Polyhedral Contact Formation
  Modeling and Identification for Autonomous Compliant Motion},''
  \emph{Robotics and Automation, IEEE Transactions on}, vol.~19, pp. 26 -- 41,
  2 2003.

\bibitem{Meeussen2008}
W.~Meeussen, J.~Rutgeerts, K.~Gadeyne, H.~Bruyninckx, and J.~De~Schutter,
  ``{Contact state segmentation using particle filters for programming by human
  demonstration in compliant motion tasks},'' \emph{Springer Tracts in Advanced
  Robotics}, vol.~39, pp. 3--12, 2008.

\bibitem{Subramani2018}
\BIBentryALTinterwordspacing
G.~Subramani, M.~Zinn, and M.~Gleicher, ``{Inferring geometric constraints in
  human demonstrations},'' \emph{Proceedings of the 2nd Annual Conference on
  Robot Learning}, no. CoRL, pp. 1--14, 2018. [Online]. Available:
  \url{http://arxiv.org/abs/1810.00140}
\BIBentrySTDinterwordspacing

\bibitem{SubramaniRAL2018}
\BIBentryALTinterwordspacing
G.~Subramani, M.~Gleicher, and M.~Zinn, ``{Recognizing Geometric Constraints in
  Human Demonstrations using Force and Position Signals},'' \emph{IEEE Robotics
  and Automation Letters}, vol. 3766, no.~c, pp. 1--1, 2018. [Online].
  Available: \url{http://ieeexplore.ieee.org/document/8264714/}
\BIBentrySTDinterwordspacing

\bibitem{goldstein2002classical}
\BIBentryALTinterwordspacing
H.~Goldstein, C.~P. Poole, and J.~L. Safko, \emph{{Classical Mechanics}}.\hskip
  1em plus 0.5em minus 0.4em\relax Addison Wesley, 2002. [Online]. Available:
  \url{https://books.google.com/books?id=tJCuQgAACAAJ}
\BIBentrySTDinterwordspacing

\bibitem{murray1994mathematical}
R.~M. Murray, Z.~Li, S.~S. Sastry, and S.~S. Sastry, \emph{{A mathematical
  introduction to robotic manipulation}}.\hskip 1em plus 0.5em minus
  0.4em\relax CRC press, 1994.

\bibitem{haug1989computer}
E.~J. Haug, \emph{{Computer aided kinematics and dynamics of mechanical
  systems}}.\hskip 1em plus 0.5em minus 0.4em\relax Allyn and Bacon Boston,
  1989, vol.~1.

\bibitem{choset2005principles}
H.~M. Choset, S.~Hutchinson, K.~M. Lynch, G.~Kantor, W.~Burgard, L.~E. Kavraki,
  and S.~Thrun, \emph{{Principles of robot motion: theory, algorithms, and
  implementation}}.\hskip 1em plus 0.5em minus 0.4em\relax MIT press, 2005.

\bibitem{lanczos2012variational}
C.~Lanczos, \emph{{The variational principles of mechanics}}.\hskip 1em plus
  0.5em minus 0.4em\relax Courier Corporation, 2012.

\bibitem{burrus2012iterative}
C.~S. Burrus, ``{Iterative reweighted least squares},'' \emph{OpenStax CNX.
  Available online: http://cnx.
  org/contents/92b90377-2b34-49e4-b26f-7fe572db78a1}, vol.~12, 2012.

\bibitem{fletcher2013practical}
R.~Fletcher, \emph{{Practical methods of optimization}}.\hskip 1em plus 0.5em
  minus 0.4em\relax John Wiley {\&} Sons, 2013.

\bibitem{Niekum2015}
S.~Niekum, S.~Osentoski, C.~G. Atkeson, and A.~G. Barto, ``{Online Bayesian
  changepoint detection for articulated motion models},'' \emph{Proceedings -
  IEEE International Conference on Robotics and Automation}, vol. 2015-June,
  no. June, pp. 1468--1475, 2015.

\bibitem{akaikeh1974}
H.~Akaike, ``{A new look at the statistical model identification},'' \emph{IEEE
  Transactions on Automatic Control}, vol.~19, no.~6, pp. 716--723, 12 1974.

\end{thebibliography}
\bibliographystyle{IEEEtran}
% biography section
% 
% If you have an EPS/PDF photo (graphicx package needed) extra braces are
% needed around the contents of the optional argument to biography to prevent
% the LaTeX parser from getting confused when it sees the complicated
% \includegraphics command within an optional argument. (You could create
% your own custom macro containing the \includegraphics command to make things
% simpler here.)
%\begin{IEEEbiography}[{\includegraphics[width=1in,height=1.25in,clip,keepaspectratio]{mshell}}]{Michael Shell}
% or if you just want to reserve a space for a photo:

\begin{IEEEbiography}[{\includegraphics[width=1in,height=1.25in,clip,keepaspectratio]{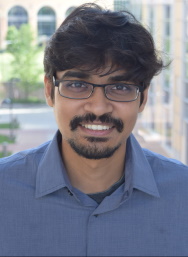}}]{Guru Subramani}
received the B.S. degree in mechanical engineering from National Institute of Technology Tiruchirappalli in 2012 and the M.S. and Ph.D. degrees in mechanical engineering from University of Wisconsin-Madison in 2014 and 2019, respectively. 

\end{IEEEbiography}

% if you will not have a photo at all:
\begin{IEEEbiography}[{\includegraphics[width=1in,height=1.25in,clip,keepaspectratio]{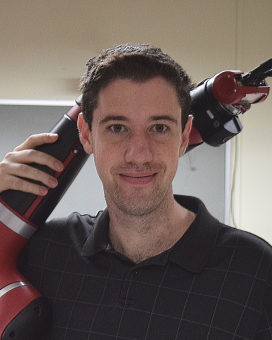}}]{Michael Hagenow}
received the B.S. degree in mechanical engineering from Tufts University in 2014 and the M.S. degree in mechanical engineering from University of Wisconsin-Madison in 2019, where he is working towards the Ph.D. degree in mechanical engineering. His research interests include robotics, control, and dynamics.
\end{IEEEbiography}

% insert where needed to balance the two columns on the last page with
% biographies
%\newpage

\begin{IEEEbiography}[{\includegraphics[width=1in,height=1.25in,clip,keepaspectratio]{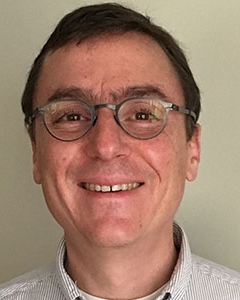}}]{Michael Gleicher} is a Professor in the Department of Computer Sciences at the University of Wisconsin, Madison.  Prof. Gleicher is founder of the Department's Visual Computing Group. His research interests span the range of visual computing, including data visualization, robotics, image and video processing tools, virtual reality, and character animation. His current foci are human data interaction and human robot interaction. Prior to joining the university, Prof. Gleicher was a researcher at The Autodesk Vision Technology Center and in Apple Computer's Advanced Technology Group. He earned his Ph. D. in Computer Science from Carnegie Mellon University, and holds a B.S.E. in Electrical Engineering from Duke University. In 2013-2014, he was a visiting researcher at INRIA Rhone-Alpes. Prof. Gleicher is an ACM Distinguished Scientist.
\end{IEEEbiography}

\begin{IEEEbiography}[{\includegraphics[width=1in,height=1.25in,clip,keepaspectratio]{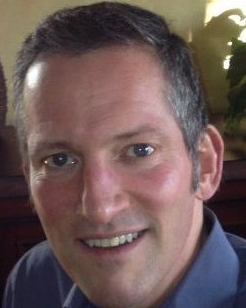}}]{Michael Zinn} (M’07) received the B.S. and
M.S. degrees from Massachusetts Institute of
Technology, Cambridge, MA, USA, in 1987 and
1988, respectively, and the Ph.D. degree in mechanical engineering from Stanford University,
Stanford, CA, USA, in 2005.
He joined the faculty at the University of Wisconsin, Madison, WI, USA, in 2007. Prior to joining the University of Wisconsin, Madison faculty,
he was the Director of Systems and Controls Engineering at Hansen Medical where he helped to
develop the world’s first commercially available minimally invasive flexible surgical robotic system. In addition to his experience at Hansen
Medical, he has more than ten years of electro-mechanical system design and manufacturing experience in aerospace and high-technology
industries. His research interests include understanding and overcoming the design and control challenges of complex electro-mechanical
systems with a primary focus on human-centered robotics. His focus
on human-centered robotics spans multiple application areas including
manufacturing, medical devices, and haptics.
\end{IEEEbiography}

% You can push biographies down or up by placing
% a \vfill before or after them. The appropriate
% use of \vfill depends on what kind of text is
% on the last page and whether or not the columns
% are being equalized.

%\vfill

% Can be used to pull up biographies so that the bottom of the last one
% is flush with the other column.
%\enlargethispage{-5in}

% that's all folks
\end{document}